\documentclass[letterpaper, 10 pt, conference]{ieeeconf}  

\IEEEoverridecommandlockouts 

\usepackage{graphicx} 
\graphicspath{{figs/}}
\usepackage{amsmath} 
\usepackage{amssymb}  
\usepackage{mathtools}

\usepackage{multicol}
\usepackage{placeins}
\usepackage{diagbox}
\usepackage{todonotes}
\usepackage{adjustbox}
\usepackage{verbatim}
\usepackage{cite}
\usepackage{wrapfig}
\usepackage{booktabs}

\usepackage[noline,linesnumbered]{algorithm2e}

\newcommand{\vect}[1]{\boldsymbol{#1}}

\newcommand{\vecto}[0]{\textrm{vec}}

\newcommand{\ourfigfont}[1]{\smaller{#1}}

\title{\LARGE Hybrid control trajectory optimization under uncertainty}

\author{Joni Pajarinen$^{1}$, Ville Kyrki$^{2}$, Michael Koval$^{3}$, Siddhartha Srinivasa$^{4}$, Jan Peters$^{5}$, and Gerhard Neumann$^{6}$ 
\thanks{This work is supported by EU Horizon 2020 project RoMaNS, project reference \#645582.}%
\thanks{$^{1}$J.~Pajarinen is with the Computational Learning for Autonomous Systems (CLAS) and Intelligent Autonomous Systems (IAS) labs, TU Darmstadt, Germany
{\tt\small pajarinen@ias.tu-darmstadt.de}}%
\thanks{$^{2}$V.~Kyrki is with the Department of Electrical Engineering and Automation, Aalto University, Finland
{\tt\small Ville.Kyrki@aalto.fi}}%
\thanks{$^{3}$M.~Koval and $^{4}$S.~Srinivasa are with The Robotics Institute, Carnegie Mellon University, USA
{\tt\small \{mkoval,siddh\}@cs.cmu.edu}}%
\thanks{$^{5}$J.~Peters is with the IAS lab, TU Darmstadt, Germany and the
  Max Planck Institute for Intelligent Systems, Tuebingen, Germany
{\tt\small peters@ias.tu-darmstadt.de}}%
\thanks{$^{6}$G.~Neumann is with the Lincoln Centre for Autonomous Systems, University of Lincoln, UK
{\tt\small neumann@ias.tu-darmstadt.de}}%
}

\begin{document}

\maketitle
\thispagestyle{empty}
\pagestyle{empty}

\begin{abstract}
  Trajectory optimization is a fundamental problem in robotics. While
  optimization of continuous control trajectories is well developed,
  many applications require both discrete and continuous, i.e.\ hybrid
  controls. Finding an optimal sequence of hybrid controls is
  challenging due to the exponential explosion of discrete control
  combinations. Our method, based on Differential Dynamic Programming
  (DDP), circumvents this problem by incorporating discrete actions
  inside DDP: we first optimize continuous mixtures of discrete
  actions, and, subsequently force the mixtures into fully discrete
  actions. Moreover, we show how our approach can be extended to
  partially observable Markov decision processes (POMDPs) for
  trajectory planning under uncertainty. We validate the approach in a
  car driving problem where the robot has to switch discrete gears and
  in a box pushing application where the robot can switch the side of
  the box to push. The pose and the friction parameters of the pushed
  box are initially unknown and only indirectly observable.
\end{abstract}

\section{INTRODUCTION}

Many control applications require both discrete and continuous, i.e.\
hybrid controls. For example, consider a car with continuous
acceleration and direction control but discrete gears
\cite{kirches09}. Switching gears changes the dynamics of the
car. Another example is pushing a box
\cite{dogar10,dogar12,koval14} where the agent can select not only a continuous
pushing direction and velocity but also a discrete side to push. Hybrid
control is important also in other applications, for example, chemical
engineering processes involving on-off valves \cite{kawajiri08}.

Hybrid control is an active research topic \cite{branicky98,bemporad00,sager05,nandola08,azhmyakov09,kirches09,zhu15}. In this paper,
we investigate systems with non-linear dynamics and long sequences of
hybrid controls and states (trajectories). We provide hybrid
control trajectory planning methods for optimizing linear feedback controllers
in systems with stochastic dynamics and partial observability
which is a challenging but common setting in robotic applications.

In the box pushing application that motivated this paper, we
investigate the problem of a robot pushing an unknown object to a
predefined location.
\begin{figure}[t]
  \centering \vspace{0.5em}
  \begin{tabular}{cc}
    \fbox{
      \begin{tabular}{c}
        \includegraphics[height=0.12\textwidth]{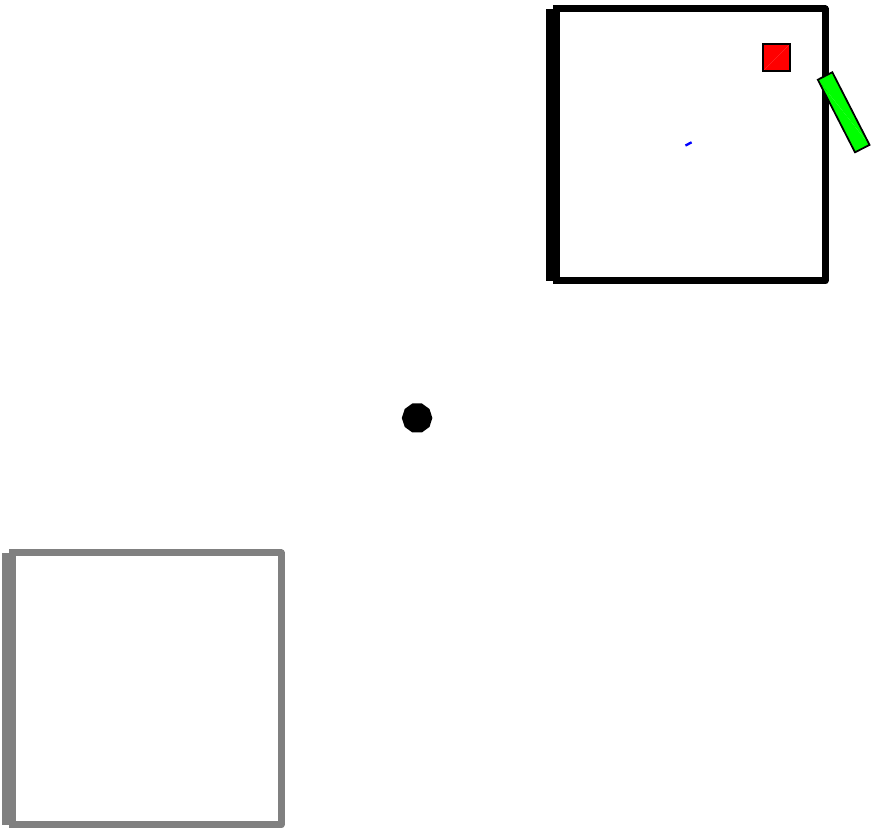}\\\hline
        \includegraphics[height=0.12\textwidth]{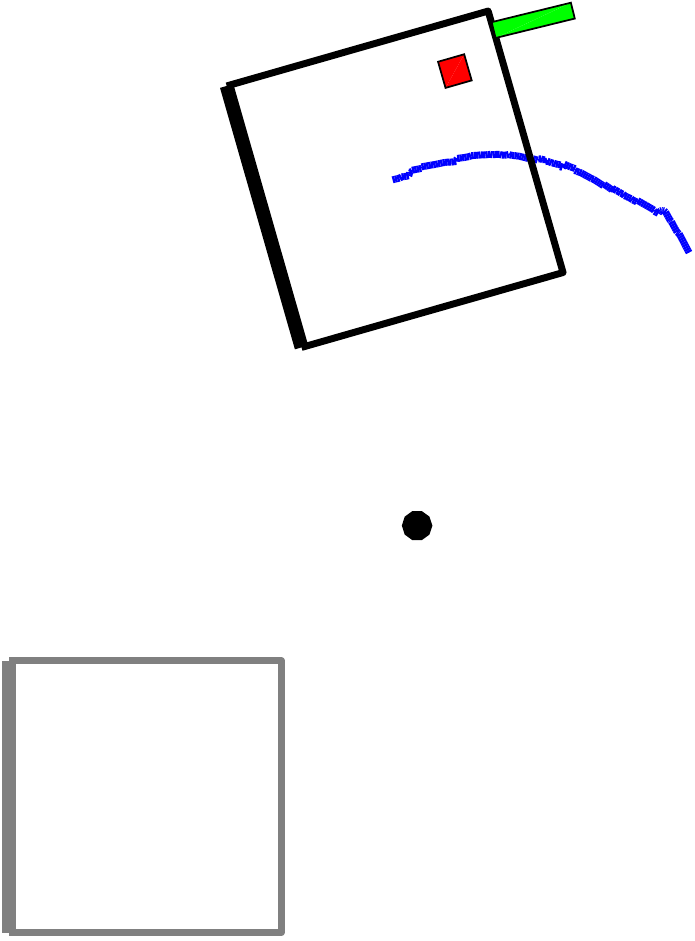}\\\hline
        \includegraphics[height=0.12\textwidth]{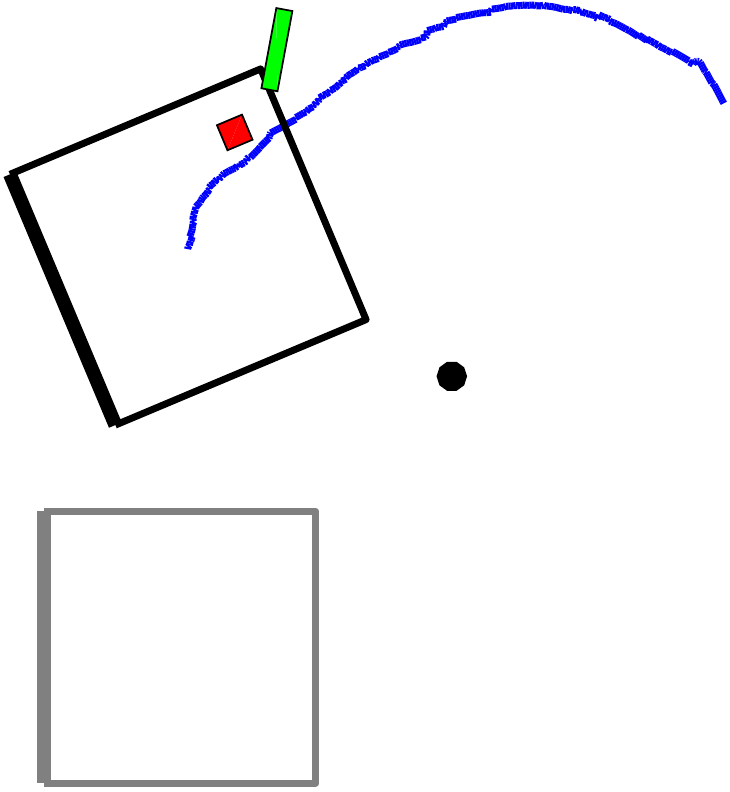}\\\hline
        \includegraphics[height=0.12\textwidth]{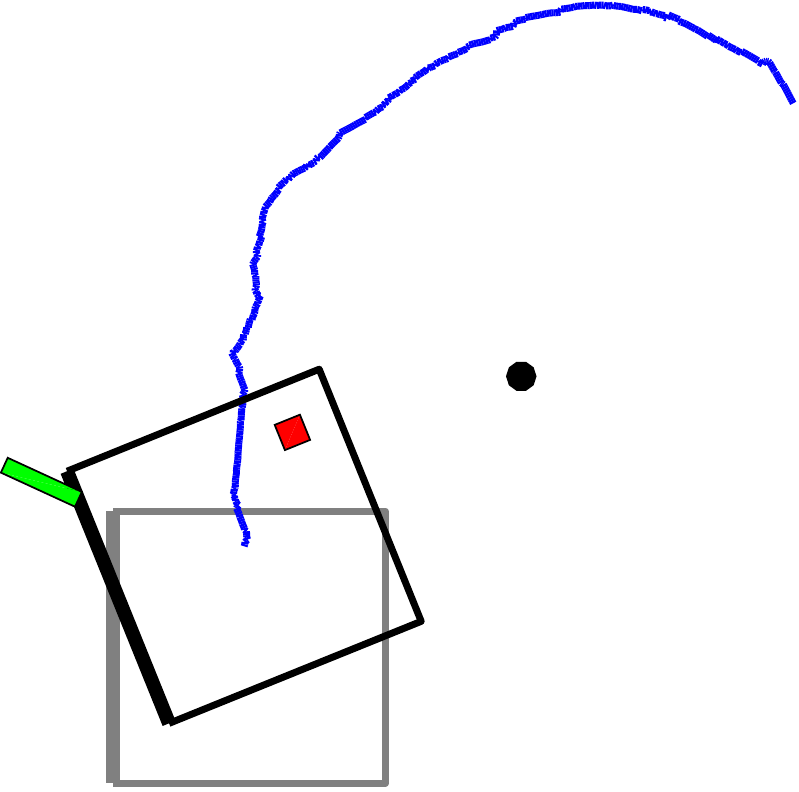}
    \end{tabular}}&
    \fbox{
      \begin{tabular}{c}
        \includegraphics[height=0.12\textwidth]{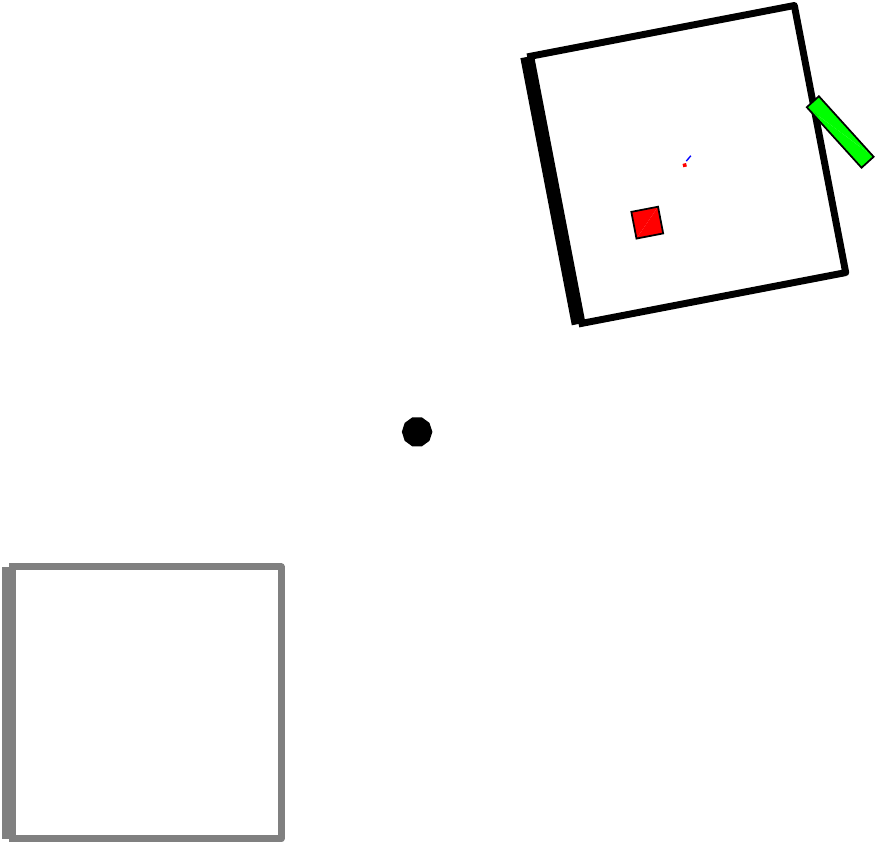}\\\hline
        \includegraphics[height=0.12\textwidth]{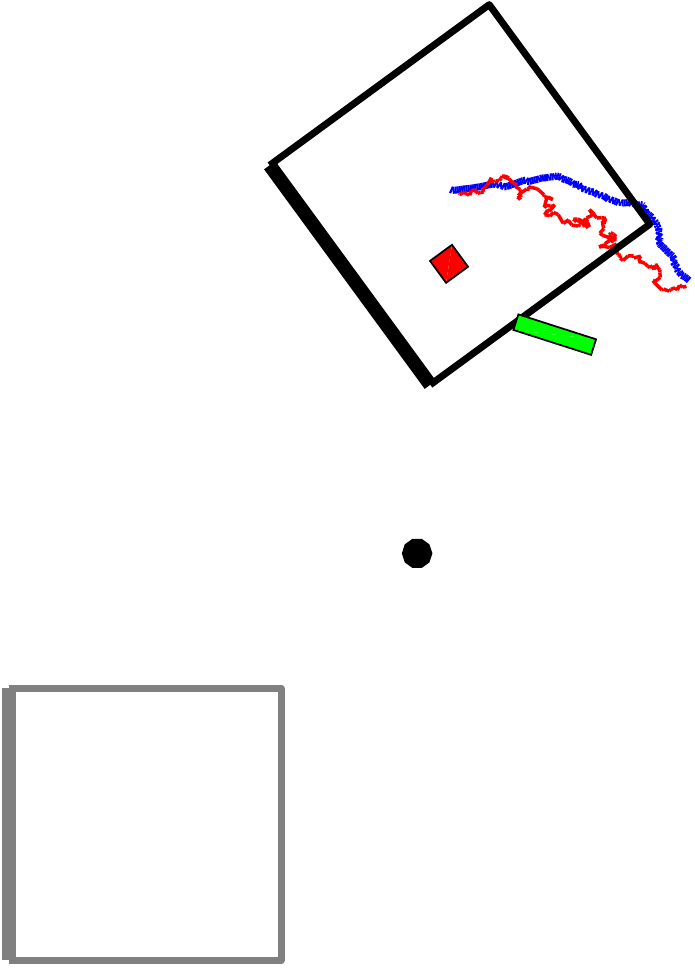}\\\hline
        \includegraphics[height=0.12\textwidth]{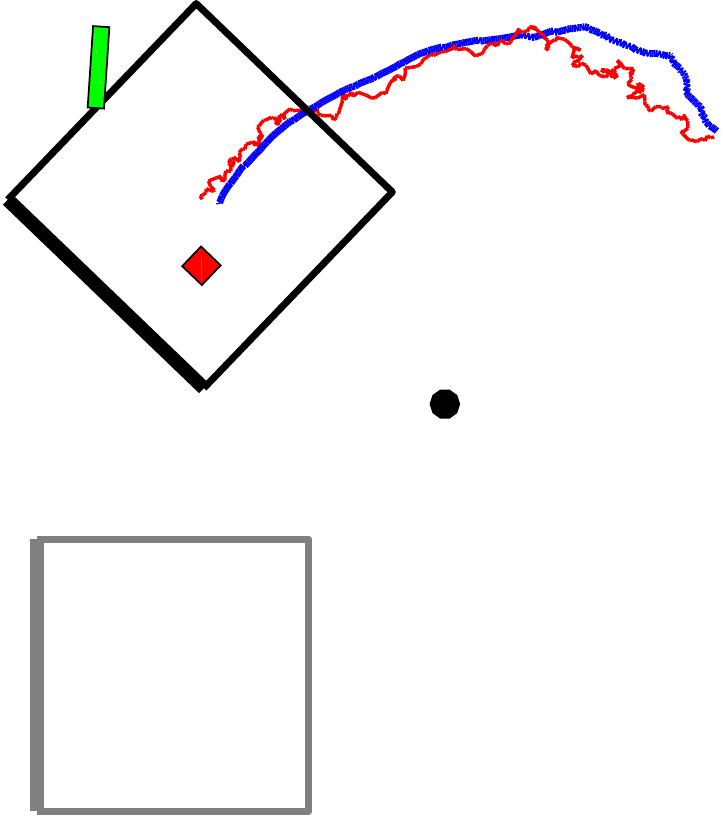}\\\hline
        \includegraphics[height=0.12\textwidth]{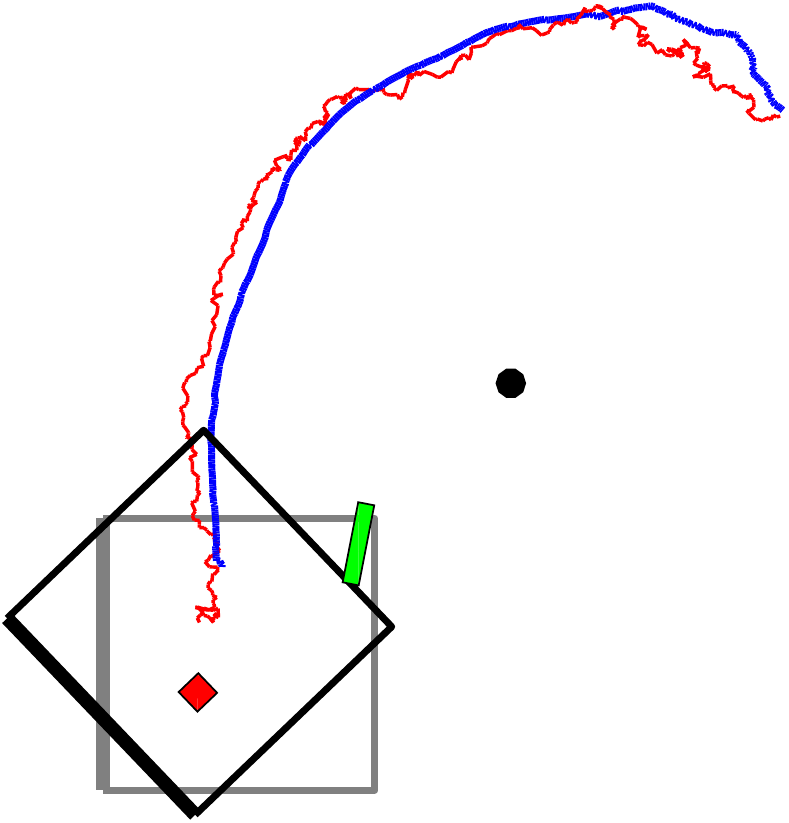}
    \end{tabular}}
  \end{tabular}
  \caption{Hybrid control box pushing. The robot tries to push the box
    with a green finger to the grey target position without hitting
    the black obstacle in the middle. The robot chooses at each time
    step a box side to push (discrete action), and a continuous
    pushing velocity and direction (continuous action). When pushed
    forward the box also rotates around the red center of
    friction. \textbf{Left}: The robot observes the box fully and can
    predict box movement. \textbf{Right}: The robot pushes a partially
    observed box with stochastic dynamics; the actual red trajectory
    differs from the expected blue trajectory. Partial observability
    makes trajectory optimization hard: pushing close to a box corner
    increases the probability of missing the box. A hybrid approach
    can directly switch the side to push and avoid \emph{considering}
    moving the finger over corners.}
  \vspace{-2em}
  \label{fig:teaser}
\end{figure}
Fig.~\ref{fig:teaser} shows examples of pushing trajectories generated
by our hybrid control approach. In object pushing, the robot may not
know in advance the physical properties of the object such as the
friction parameters or the center of friction which is the point
around which the object rotates when pushed. The actual center of
friction and friction parameters may vary considerably between
different objects. Moreover, the robot can only make noisy
observations about the current pose of the object, using, for example,
a vision sensor. The robot needs to take this observation and dynamics
uncertainty into account in order to accomplish its task. For example,
when pushing a box to a predefined location, the robot needs to
consider how to move its finger along the box edge. However, if the
robot is not certain about the actual pose of the box it may miss the
box when pushing close to the box corners. This may make approaches
with continuous control, such as differential dynamic programming
(DDP) reluctant to \emph{consider} moving the finger around box
corners and converge to local optima as shown experimentally in
Section~\ref{sec:experiments}. A hybrid approach can directly switch
the side of the box to push and succeed under uncertainty.

For modeling trajectory optimization under both uncertain dynamics and
sensing we use a partially observable Markov decision process (POMDP),
and, model uncertain box pushing parameters as part of the POMDP state
space. Moreover, control limits are common in robotics; for example,
joint motors have physical limits. In the pushing application, the
pushing velocity and direction are limited. We show how to add hard
limits to DDP based POMDP trajectory optimization, which has so far
only been shown for deterministic MDPs
\cite{tassa14}. Below, we summarize the major contributions of this
paper:
\begin{itemize}
  \item \textbf{Hybrid Trajectory Optimization:} We introduce
    trajectory optimization with discrete and continuous actions using
    differential dynamic programming (DDP).
  \item \textbf{Extension to Partially Observable Environments:} We
    introduce the first POMDP trajectory optimization algorithm with
    hybrid controls.
  \item \textbf{Hard Control Limits:} Inspired by hard-control limits
    for trajectory optimization \cite{tassa14}, we introduce hard control
    limits for POMDP trajectory optimization.
  \item \textbf{Box-Pushing Application:} We use for the first time a
    POMDP formulation for pushing an unknown object in a task specific
    way. Taking uncertainty into account is required for performing
    the task.
\end{itemize}

\section{RELATED WORK}

In this paper, we consider hybrid control
\cite{branicky98,riedinger99,bemporad00,sager05,nandola08,azhmyakov09,kirches09,zhu15}
for finite discrete time trajectory optimization \cite{vonstryk92} in
systems with non-linear stochastic dynamics and noisy, partial state
information. We optimize time varying linear feedback controllers producing a
trajectory consisting of states, covariances, and hybrid controls.

There are many methods for optimizing continuous control trajectories
\cite{vonstryk92,platt10,van12,patil15}. Discrete controls present a
challenge due to the exponential number of discrete control
combinations w.r.t.\ the planning horizon. The naive approach of
optimizing continuous controls for each combination of discrete
controls scales only to a few time steps. \cite{lincoln02} uses a tree
of linear quadratic regulator (LQR) solutions for discrete action
combinations and introduces a technique for pruning the tree but the
tree may still grow exponentially over time.

\cite{zhu15} tries to find a local optimum by performing continuous
control optimization and local discrete control improvements
iteratively. Hybrid control problems can be also translated into mixed
integer non-linear programming (MINLP) problems. However, in complex
problems, hybrid control MINLP solutions can be restricted to only a
few time step horizons~\cite{nandola08}. \cite{kirches09} shows how to
apply ``convexification'', introduced in \cite{sager05}, for discrete
MINLP variables in fully observable deterministic dynamic
problems. ``Convexification'' transforms discrete controls into
weights replacing the original dynamics function with a convex
mixture. We show how to apply a similar idea to differential dynamic
programming (DDP) with stochastic dynamics and partial
observations. Instead of using hybrid control for optimizing
trajectories, reinforcement learning approaches based on the options
framework can compute high level discrete actions, also called options
\cite{daniel16}, and execute a continuous control
policy for each high level action.

\textbf{RRTs.} Rapidly-exploring random trees (RRTs) \cite{lavalle01}
are often used for trajectory initialization. \cite{branicky02} uses
RRTs in hybrid control trajectory planning of fully observable
systems. \cite{zito12} uses also hybrid, that is, discrete and
continuous actions, for pushing objects from an initial three
dimensional configuration into another final configuration. Contrary
to our work, \cite{zito12} does not do trajectory optimization and
does not take uncertainty into account.

\textbf{POMDPs.} Our proposed trajectory optimization approach plans
under model, sensing, and actuation uncertainty. \cite{egerstedt06}
and \cite{azuma06} investigate partial observations for computing
switching times in switched systems by providing simple analytic
examples of a few time steps. However, \cite{egerstedt06,azuma06} do
not model state uncertainty. Previously, planning under uncertainty
has been investigated in simulated robotic tasks in
\cite{ross08,dallaire09,bai14}. \cite{bai14} presents a sampling
based POMDP approach for continuous states and actions. \cite{bai14}
uses the POMDP approach for Bayesian reinforcement learning in a
simulated pendulum swingup experiment.
\cite{agha14} use feedback based motion-planning with
uncertainty and partial observations. \cite{agha14} target
navigation type of applications and use probabilistic roadmap planning
to generate a graph where graph nodes represent locations.
One classic way of dealing with observation uncertainty is to assume
maximum likelihood observations~\cite{platt10}. \cite{patil15} uses
shooting methods for POMDP trajectory optimization. The POMDP approach
of \cite{van12} is based on iterated linear quadratic Gaussian (iLQG)
control with covariance linearization. In this paper, we extend the
fully observable iLQG/DDP \cite{tassa14} algorithm as well as the
iLQG/DDP based POMDP approach of \cite{van12} to hybrid controls. We
also show a straightforward way of using hard control limits with
iLQG/DDP based POMDP.

\textbf{Box pushing.} When pushing an unknown object a robot needs to
plan its motions and pushing actions while taking
model~\cite{kopicki16}, sensing, and actuation uncertainty into
account. Therefore, we model the pushing task as a hybrid control
continuous state POMDP. We base our box pushing simulation on the same
quasi-static physics model \cite{lynch92} utilized in
\cite{dogar10,dogar12,koval14}. \cite{lynch92} shows how to find out
friction parameters but not how to plan for a specific
task. \cite{koval16} discretizes the state space potentially
increasing the state space size exponentially w.r.t.\ the number of
dimensions (also known as the state-space explosion problem). Instead
of prespecified pushing motions, our approach could be used for
planning pushing trajectories in the higher level task planning
approach of~\cite{dogar10,dogar12} to handle partly known objects or a
specific task.

\section{PRELIMINARIES}
\label{sec:hybrid_trajectory_optimization}

In this section, we first define the problem and subsequently discuss
differential dynamic programming, which we extend to hybrid control in
the following sections.
\begin{figure}[t]
  \centering
  \vspace{0.5em}
  \includegraphics[width=0.4\textwidth]{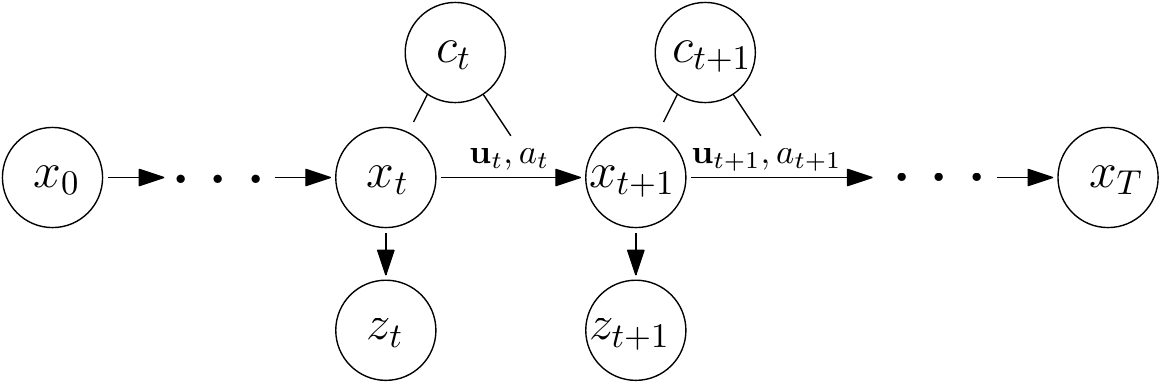}
  \caption{Graphical illustration of hybrid sequential control. At time step
    $t$ the agent executes continuous and discrete controls
    $\vect{u}_t$ and $a_t$, pays a cost $c_t(\vect{x}_t, \vect{u}_t,
    a_t)$, and the state changes from $\vect{x}_t$ to
    $\vect{x}_{t+1}$. Under partial observability the agent does not
    see $\vect{x}_t$ but instead makes an observation
    $\vect{z}_t$. The goal is to minimize the total cost over a finite
    number of time steps.}
  \vspace{-1em}
  \label{fig:graphical_model}
\end{figure}
\subsection{Problem statement}
In finite time-discrete hybrid sequential control, at each time step
$t$, the agent executes continuous control $\vect{u}_t$\footnote{In
  reinforcement learning and AI research $s$ often denotes the state,
  $a$ the control or action, and rewards, corresponding to negative
  costs, are used for specifying the optimization goal.} and discrete
control $a_t$ out of $N_a$ possibilities, incurs a cost of the form
$c_t(\vect{x}_t, \vect{u}_t, a_t)$ for intermediate time steps and
traditionally $c_T(\vect{x}_t)$ for the final time step
$T$. Subsequently, the world changes from state $\vect{x}_t$ to state
$\vect{x}_{t+1}$ according to a dynamics function $\vect{x}_{t+1} =
\vect{f}(\vect{x}_t, \vect{u}_t, a_t)$. The goal is to minimize the
total cost $c(\vect{x}_T)+\sum_{t=0}^{T-1}
c(\vect{x}_t,\vect{u}_t,a_t)$. Fig.~\ref{fig:graphical_model}
illustrates this process.

In a partially observable Markov decision process (POMDP), the agent
does not directly observe the system state but instead makes an
observation $\vect{z}_{t}$ at each time step $t$. In a POMDP, state
dynamics and the observation function are stochastic. While the agent
does not observe the current state directly, the agent can choose the
control based on the belief $P(\vect{x}_t)$, a probability
distribution over states, which summarizes past controls and
observations and is a sufficient statistic for optimal decision
making.

Because solving POMDPs exactly is intractable even for small discrete
problems, Gaussian belief space planning methods
\cite{van12,webb14,patil15} assume a multi-variate Gaussian
distribution for the state space, transitions, and observations with a
Gaussian initial belief distribution $\vect{x}_0 \sim
\mathcal{N}(\hat{\vect{x}}_0, \vect{\Sigma}_0)$. We denote with
$\hat{\vect{x}}$ the mean and with $\vect{\Sigma}$ the covariance of the belief. In a
hybrid control Gaussian POMDP, the next state depends on a possibly
non-linear function $\vect{f}(\vect{x}_t, \vect{u}_t, a_t)$:
\begin{equation}
  \vect{x}_{t+1} = \vect{f}(\vect{x}_t, \vect{u}_t, a_t) + \vect{m} \;,
  \label{eq:transition}
\end{equation}
where $\vect{m} \sim \mathcal{N}(0, M(\vect{x}, \vect{u}, a))$ is state
and control specific multi-variate Gaussian noise. The observation
$\vect{z}_t$ at each time step is specified by a possibly non-linear
function $\vect{h}(\vect{x}_t)$ with additive Gaussian noise:
\begin{equation}
  \vect{z}_t = \vect{h}(\vect{x}_t) + \vect{n} \;,
  \label{eq:observation}
\end{equation}
where $\vect{n} \sim \mathcal{N}(0, N(\vect{x}))$ is state specific
multi-variate Gaussian noise.

The immediate cost function is of the form $c(\vect{x}, \vect{u}, a,
\vect{\Sigma})$ for intermediate time steps and $c(\vect{x})$ for the
final time step. Many methods use the covariance as a term in their
cost function to penalize high uncertainty in state estimates.

\subsection{Differential dynamic programming}
\label{sec:DDP}

DDP~\cite{mayne66} is a widely used method for trajectory optimization
with fast convergence \cite{jacobson70,liao92} and the ability to
generate feedback controllers. In this section, we discuss DDP and
iterative linear quadratic Gaussian (iLQG) \cite{todorov05}, a special
version of DDP which disregards second order dynamics derivatives and
adds regularization and line search to deal with non-linear dynamics.

We will describe DDP here briefly. Please, see \cite{tassa14} for a
more detailed description. DDP optimization starts from an initial
nominal trajectory, a sequence of controls and states and then applies
back and forward passes in succession. In the backwards pass, DDP
quadratizes the value function around the nominal trajectory and uses
dynamic programming to compute linear forward and feedback gains at
each time step. In the forward pass, DDP uses the new policy to
project a new trajectory of states and controls. The new trajectory is
then used as nominal trajectory in the next backwards pass and so
on. A short description follows but please see
\cite{mayne66,todorov05,tassa14} for more details.

Denote the nominal trajectory with upper bars and the differences
between the state and control w.r.t.\ the nominal trajectory as
$\Delta \vect{x} = \vect{x} - \vect{\bar{x}}$ and $\Delta \vect{u} =
\vect{u} - \vect{\bar{u}}$, respectively. DDP assumes that the value
function at time step $t$ is of quadratic form
\begin{equation}
  V_t(\vect{x}) = V + \Delta \vect{x}^T \vect{V}_{\vect{x}} + 
  \Delta \vect{x}^T \vect{V}_{\vect{x}\vect{x}} \Delta \vect{x}.
  \label{eq:value_function},
\end{equation}

The one time step value difference between time step $t$ and $t+1$
w.r.t.\ states and controls
\begin{align*}
  Q_t(\Delta \vect{x}, \Delta \vect{u}) =
  c_t(\vect{x} + \Delta \vect{x}, \vect{u} + \Delta \vect{u}) -
  c_t(\vect{x}, \vect{u}) + \\
  V_{t+1}(\vect{f}(\vect{x} + \Delta \vect{x},
                  \vect{u} + \Delta \vect{u})) -
  V_{t+1}(\vect{f}(\vect{x}, \vect{u}))
\end{align*}
is also assumed quadratic.

Given $V_{t+1}(\vect{x})$ we can compute the continuous control
$\vect{u}$ at time step $t$:
\begin{align}
  \vect{u} &= \vect{K} (\Delta \vect{x}) + \vect{k} + \vect{\bar{u}} 
  \label{eq:delta_u} \\
  \vect{K} &= - \vect{Q}_{\vect{u}\vect{u}}^{-1} \vect{Q}_{\vect{u}\vect{x}} \;\;\;\;\textrm{and}\;\;\;\;
  \vect{k} = - \vect{Q}_{\vect{u}\vect{u}}^{-1} \vect{Q}_{\vect{u}} \;. \label{eq:kK}
\end{align}
where $\vect{K}$ is the feedback gain and $\vect{k}$ the forward gain
of the linear feedback controller. $\vect{Q}_{\vect{u}\vect{u}}$,
$\vect{Q}_{\vect{u}\vect{x}}$, and $\vect{Q}_{\vect{u}}$ are computed
based on $Q_t(\Delta \vect{x}, \Delta \vect{u})$ as detailed
in~\cite{tassa14}.

Recently, \cite{tassa14} introduced a version of DDP that allows
efficient computation with hard control limits. Hard control limits
require quadratic programming (QP) for computing the forward gains
$\vect{k}$:
\begin{align}
  \vect{k} = \arg\min_{\Delta\vect{u}} \frac{1}{2} \Delta\vect{u}^T
  \vect{Q}_{\vect{u}\vect{u}} \Delta\vect{u} + \Delta\vect{u}^T \vect{Q}_{\vect{u}} \nonumber \\
  \vect{u}_{\textrm{LB}} \le \vect{u} + \Delta\vect{u} \le \vect{u}_{\textrm{UB}} \;, \label{eq:QP}
\end{align}
where $\vect{u}_{\textrm{LB}}$ and $\vect{u}_{\textrm{UB}}$ are the
lower and upper limits, respectively. For the feedback gain matrix
$\vect{K}$, the rows corresponding to clamped controls are set to
zero. To solve QPs, \cite{tassa14} provides a gradient descent
algorithm which allows initialization of the QP with a previously
computed forward gain. Good initialization makes the approach
computationally efficient. In the next Section, we will discuss how to
extend the QP approach with equality constraints which allows action
probabilities needed by our hybrid control approach.

\section{HYBRID TRAJECTORY OPTIMIZATION}
\label{sec:hybrid_DDP}

The first problem with hybrid control is that the discrete action
choice depends on the combination of discrete actions at all time
steps resulting in exponentially many combinations. The second problem
is that in non-linear problems we can adjust the approximation error
due to linearization for continuous but not for discrete controls. For
example, continuous iLQG adjusts the linearization error by scaling
the forward gain $\vect{k}$ with a real valued parameter $\alpha$
during the forward pass (when $\alpha$ approaches zero the
linearization error approaches zero). However, for discrete actions,
decreasing the amount of control change is not straightforward.

Below, we present three approaches for optimizing hybrid control DDP
policies. The first two are simple greedy baseline approaches which we
provide for comparison. The third more powerful approach uses a
continuous mixture of discrete actions driving the mixture during
optimization into single selection using a special cost function.

\subsection{Greedy discrete action choice}
\label{sec:greedy_actions}
During the DDP back pass, at each time step, the greedy approach
computes the expected value at the nominal state and control for each
discrete action separately and selects the action which yields minimum
expected cost. In the greedy approach, there is no feedback control
for discrete actions, only the fixed actions computed during the DDP
back pass.

\subsection{Interpolated discrete action choice}
\label{sec:interpolated_actions}
The second baseline approach for hybrid control attempts to smoothly
scale the linearization error w.r.t.\ discrete controls. The approach
interpolates between nominal and new optimized discrete
actions. First, the interpolated approach computes new actions
identically to the greedy approach, but, then uses only a fraction
$\alpha$ of the new discrete actions which differ from the old nominal
actions. The selection of actions is done evenly over the time
steps. For example, for $\alpha=0.5$ every other new discrete control
would be used.

\subsection{Mixture of discrete actions}
\label{sec:mixed_actions}
The approach that we propose for hybrid control uses a mixture of
discrete actions assigning a continuous pseudo-probability to each
discrete action. During optimization, using a specialized cost
function which is discussed further down, we drive the mixture to
select a single discrete action.

Our modified control $\hat{\vect{u}}$ is
\begin{equation}
  \hat{\vect{u}} = \begin{bmatrix} \vect{u}\\ \vect{p_a}
  \end{bmatrix} \;,
\end{equation}
where $\vect{p_a}$ contains the action probabilities. The
dimensionality of the controls increases by the number of discrete
actions. For simplicity, we assume here a single discrete control
variable but our approach directly extends to several discrete control
variables. For several discrete controls, the dimensionality would
either be the sum of discrete control dimensions if one treats
discrete controls as independent, or, the product of discrete control
dimensions if one wants better accuracy.

For hybrid controls, the dynamics model $\vect{f}(\vect{x}, \vect{u},
a)$ depends on both continuous controls $\vect{u}$ and discrete
actions $a$. The new dynamics model is a mixture of the original one:
\begin{equation}
  \hat{\vect{f}}(\vect{x}, \hat{\vect{u}}) = 
  \sum_a p_a \vect{f}(\vect{x}, \vect{u}, a) \;.
  \label{eq:dynamics}
\end{equation}
Note that (\ref{eq:dynamics}) is essentially identical to the
``convexified'' dynamics function in \cite{sager05} for fully
observable MINLP hybrid control. However, \cite{sager05} does not
present a mixture model for immediate cost functions and our special
cost function further down that forces the system into a bang-bang
solution differs from the one in \cite{sager05} because of the
positive-definite Hessian for cost functions in DDP.

For hybrid controls, we define the cost function $c(\vect{x},
\vect{u}, a)$ in the mixture model as
\begin{equation}
  \hat{c}(\vect{x}, \hat{\vect{u}}) = 
  \sum_a \phi(p_a) c(\vect{x}, \vect{u}, a) \;,
\end{equation}
where $\phi(\cdot)$ is a smoothing function to make the Hessian of the
cost function positive-definite w.r.t.\ the linear parameters $p_a$. In the experiments, we used a
pseudo-Huber smoothing function 
\begin{equation}
  \phi(p) = \phi(p,0.01), \phi(p,k) = \sqrt{p^2 + k^2} - k
  \label{eq:pseudo_huber}
\end{equation}
which is close to linear but has a positive second derivative.

To optimize the forward and feedback gains during dynamic
programming, we add the following inequality and equality constraints
for the probabilities to the quadratic program in Equation
(\ref{eq:QP}):
\begin{equation}
  \vect{0} \le \vect{p} \le \vect{1} \;,\;\; \sum_a p_a = 1 \;.\label{eq:sum_pa}
\end{equation}
We extend the efficient gradient descent method for quadratic
programming from~\cite{tassa14} to deal with the equality
constraint~(\ref{eq:sum_pa}). Shortly: 1) we subtract the mean from
the search direction of probabilities satisfying the equality
constraint, 2) we modify the Armijo line search step size dynamically
so that we do not overstep probability inequality constraints.

$\hat{\vect{f}}(\vect{x}, \vect{u}, \vect{p})$ can be seen as the
expected dynamics of a partially stochastic policy. For such expected
dynamics there may not be any actual control values that would result
in such dynamics. For example, in the box pushing application a
mixture of discrete actions could correspond to pushing with several
fingers although the robot may only have one finger. However, allowing
for stochastic discrete actions in the beginning of optimization
allows convergence to a good solution, even if we force the actions to
become deterministic later. Our optimization procedure takes care of
the major problems with sequential decision making with hybrid
controls: the procedure allows to continuously decrease the
approximation error due to linearization making local updates possible
but is not subject to the combinatorial explosion of discrete action
combinations. Next we discuss how to force a deterministic policy for
discrete actions.

\textbf{Forcing deterministic discrete actions.} In the end, we want a
valid deterministic policy for discrete actions. Therefore, we
explicitly assign a cost to stochastic discrete actions that increases
during optimization driving stochastic discrete controls into
deterministic ones. Entropy would be a natural, widely used, cost
measure. However, the Hessian matrix of an entropy based cost function
is not positive-definite (the second derivative is always
negative). Instead, we use the following similarly shaped smoothed
piece-wise cost function on stochastic discrete actions
\begin{equation}
  c_{\textrm{ST}}(\vect{x}, \vect{u}, \vect{p}) = 
  C_{\textrm{ST}}
  \sum_a
  \begin{dcases}
    \phi(p_a) & \text{if } p_a < p_{\textrm{th}} \\
    \phi\left(\frac{(1 - p_a)}{p_{\textrm{th}} / (1 - p_{\textrm{th}})}\right) & \text{if } p_a \geq p_{\textrm{th}} \\
  \end{dcases}\;,
\end{equation}
where $p_{\textrm{th}} = 1 / N_a$ and $C_{\textrm{ST}}$ is an adaptive
constant. Note that while $c_{\textrm{ST}}(\vect{x}, \vect{u},
\vect{p})$ is discontinuous at $p_{\textrm{th}}$, derivatives can be
computed below and above $p_{\textrm{th}}$ and the cost drives
solutions away from $p_{\textrm{th}}$ for increasing
$C_{\textrm{ST}}$. When at $p_{\textrm{th}}$, which
corresponds to a uniform distribution, the cost achieves its maximum.
Note that any cost measure with zero cost for
probabilities $0$ and $1$ is bound to have a discontinuity when the
second derivative has to be positive. Intuitively, the positive second
derivative forces the graph of the cost measure to always curve
upwards resulting in a discontinuity at the point where the graph
starting from $0$ meets the graph ending at $1$.
\begin{wrapfigure}[5]{r}{0.10\textwidth}
  \vspace{-1.8em}
  \begin{center}
    \includegraphics[width=0.10\textwidth]{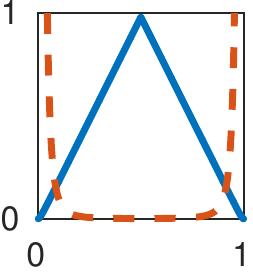}
  \end{center}
\end{wrapfigure}
Tiny image on the right shows the cost function (solid) and its second
derivative (dashed) for two discrete states and $C_{\textrm{ST}}=
1$. The x-axis shows the state probability. 

\textbf{Practicalities.} During optimization we double
$C_{\textrm{ST}}$ everytime the cost decrease between DDP iterations
is below a threshold. We set $C_{\textrm{ST}}$ to a maximum value when
half of the maximum number of iterations has elapsed. This allows for
both smoothly increasing determinicity of discrete actions and then
finally deterministic action selection. Since we run the algorithm for
a finite number of iterations and do not increase $C_{\textrm{ST}}$ to
infinity some tiny stochasticity may remain. Therefore, we select the
most likely discrete action during evaluation. Note also that due to
the linear feedback control affecting probabilities we normalize
probabilities during the forward pass.

\section{HYBRID TRAJECTORY OPTIMIZATION FOR POMDPS}
\label{sec:hybrid_POMDP}

We are now ready to discuss hybrid control of POMDP trajectories. We
start with an iLQG approach for POMDPs \cite{van12} and then discuss
how the iLQG POMDP can be extended with hybrid controls and hard
control limits.

\subsection{DDP for POMDP trajectory optimization}
\label{sec:DDP_POMDP}

The iLQG POMDP approach of~\cite{van12} extends iLQG~\cite{tassa14} to
POMDPs by using a Gaussian belief
$\mathcal{N}(\hat{\vect{x}},\vect{\Sigma})$, instead of the fully
observable state $\vect{x}$. In the forward pass iLQG POMDP uses a
standard extended Kalman filter (EKF) to compute the next time step
belief. For the backward pass, iLQG POMDP linearizes the covariance in
addition to quadratizing states and controls. The value function in
Equation~(\ref{eq:value_function}) becomes
\begin{equation}
  V_t(\hat{\vect{x}}, \vect{\Sigma}) = V + 
  \Delta \hat{\vect{x}}^T \vect{V}_{\hat{\vect{x}}} + 
  \Delta\hat{\vect{x}}^T \vect{V}_{\hat{\vect{x}}\hat{\vect{x}}} \Delta\hat{\vect{x}}
  + \vect{V}^T_{\vect{\Sigma}} \vecto[\Delta \vect{\Sigma}],
\end{equation}
where $\vecto[\Delta \vect{\Sigma}]$ is the difference between the
current and nominal covariance stacked column-wise into a vector and
$\vect{V}^T_{\vect{\Sigma}}$ is a new linear value function parameter. The related
control law/policy is shown in~\cite[Equation (23)]{van12}.

\subsection{Hybrid control for POMDP trajectory optimization}
\label{sec:hybrid_control_DDP_POMDP}
In the partially observable case, the direct and indirect cost of
uncertainty propagates through $\vect{V}_{\vect{\Sigma}}$ into other
value function components and the optimal policy has to take
uncertainty into account. However, the control law in~\cite[Equation
(23)]{van12} in iLQG POMDP is of equal form to the one in basic iLQG
shown in Equation~(\ref{eq:kK}). The only difference is that
$\vect{Q}_{\vect{u}\vect{u}}$, $\vect{Q}_{\vect{u}\vect{x}}$, and
$\vect{Q}_{\vect{u}}$ shown in Equation~(\ref{eq:kK}) are influenced
by $\vect{V}_{\vect{\Sigma}}$ from future time
steps~\cite{van12}. This means that we can directly optimize controls
using the quadratic program (QP) shown in Equation
(\ref{eq:QP}). Therefore, we can use hard limits and equality
constraints on continuous controls in iLQG POMDP which is one
technical insight in this paper.

A question this raises is whether we can also apply our proposed
hybrid control approach to iLQG POMDP? Yes. Since our method of
discrete action mixtures described in Section~\ref{sec:mixed_actions}
hides the action mixture inside the dynamics and cost functions, and,
since the observation function does not directly depend on the
controls, we can directly use the proposed hybrid control approach for
POMDP optimization.

\section{EXPERIMENTS}
\label{sec:experiments}

We experimentally validate our hybrid control approach ``Mixture'',
described in Section~\ref{sec:mixed_actions}, in two different
simulations: autonomous car driving and pushing of an unknown box. We
are not aware of previous algorithms for trajectory planning under
uncertainty which operate directly on both continuous and discrete
actions. However, in the car driving and box pushing applications, we
can reasonably map the hybrid controls directly to continuous controls
and compare against continuous iLQG~\cite{tassa14} and continuous iLQG
POMDP~\cite{van12}, extended to support hard control limits, as
described in Section~\ref{sec:hybrid_control_DDP_POMDP}. Note that in
many hybrid control applications, for example, with discrete switches
or on-off valves, mapping hybrid controls to continuous ones may not
be possible and a hybrid approach is required. We also compare with
greedy action selection ``Greedy'' described in
Section~\ref{sec:greedy_actions}, and interpolated greedy action
selection ``Interpolate'' described in
Section~\ref{sec:interpolated_actions}. We used a time horizon of
$T=500$ and ran up to $400$ optimization iterations for each
method. We set the maximum for $C_{\textrm{ST}}$ (please, see
Section~\ref{sec:mixed_actions}) to $1.28$. $C_{\textrm{ST}}$ starts
from zero, increases to $0.01$, and then doubles every time the cost
difference is below $0.01$ in box pushing and $0.0001$ in car driving.

\subsection{Autonomous car driving}
In autonomous car driving, the robot drives a nonholonomic car,
switches discrete gears, accelerates, and tries to steer the car to
zero position and pose. The dynamics are identical to the car parking
dynamics in \cite{tassa14} with a few differences. The system state
$\vect{x} = (x, y, w, v_{\textrm{CAR}})$ consists of the position of
the car $(x,y)$, the car angle $w$, and the car velocity
$v_{\textrm{CAR}}$. The continuous controls $\vect{u} =
(w_{\textrm{WHEEL}}, acc_{\textrm{CAR}})$ select the front wheel angle
$w_{\textrm{WHEEL}} \in [-0.5, 0.5]$ and car acceleration
$acc_{\textrm{CAR}} \in [0, 0.5]$. We have three discrete actions: the
robot can select either 1st or 2nd gear, or alternatively break.

In 2nd gear acceleration is halved and for breaking acceleration is
negative. The break, 1st and 2nd gear have a soft velocity limit: for
the break and 2nd gear, when $v_{\textrm{CAR}} > 4$, and for the 1st
gear when $v_{\textrm{CAR}} > 1$, the car is assigned a negative
acceleration of $acc_{\textrm{CAR}} = -0.1$ simulating real world
engine breaking. Since the 1st gear has a lower soft velocity limit
than the 2nd gear but higher acceleration, to achieve high speeds
quickly one needs to accelerate first with the 1st gear and then
switch to the 2nd gear.

\begin{figure}
  \vspace{0.3em}
  \centering
  \includegraphics[width=7cm]{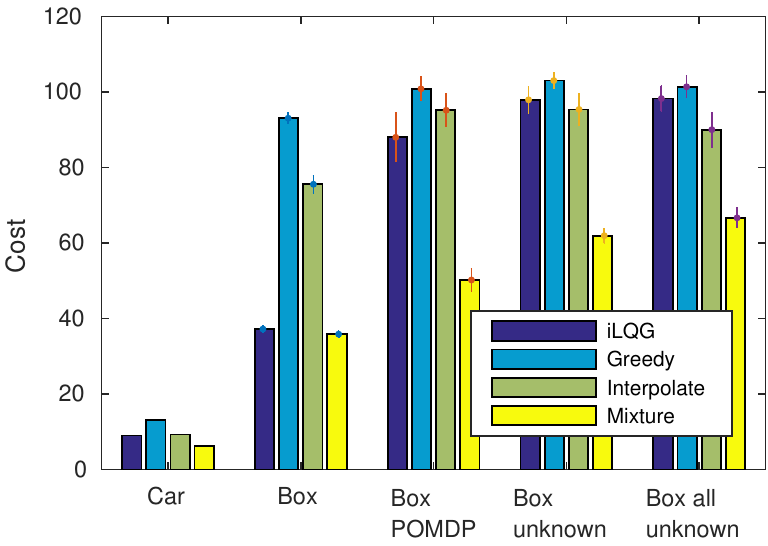}\vspace{-0.5em}
  \caption{Mean costs and standard errors of the mean for the
    comparison methods. The deterministic ``Car'' problem has a single
    trajectory and thus no standard error. ``Mixture'' performs
    overall best and is better than continuous ``iLQG'' in the three
    stochastic problems.}
  \vspace{-2em}
  \label{fig:results}
\end{figure}

\begin{figure*}
  \centering
  \vspace{0.4em}
  \begin{tabular}{cccc}
    \fbox{\includegraphics[width=3.2cm]{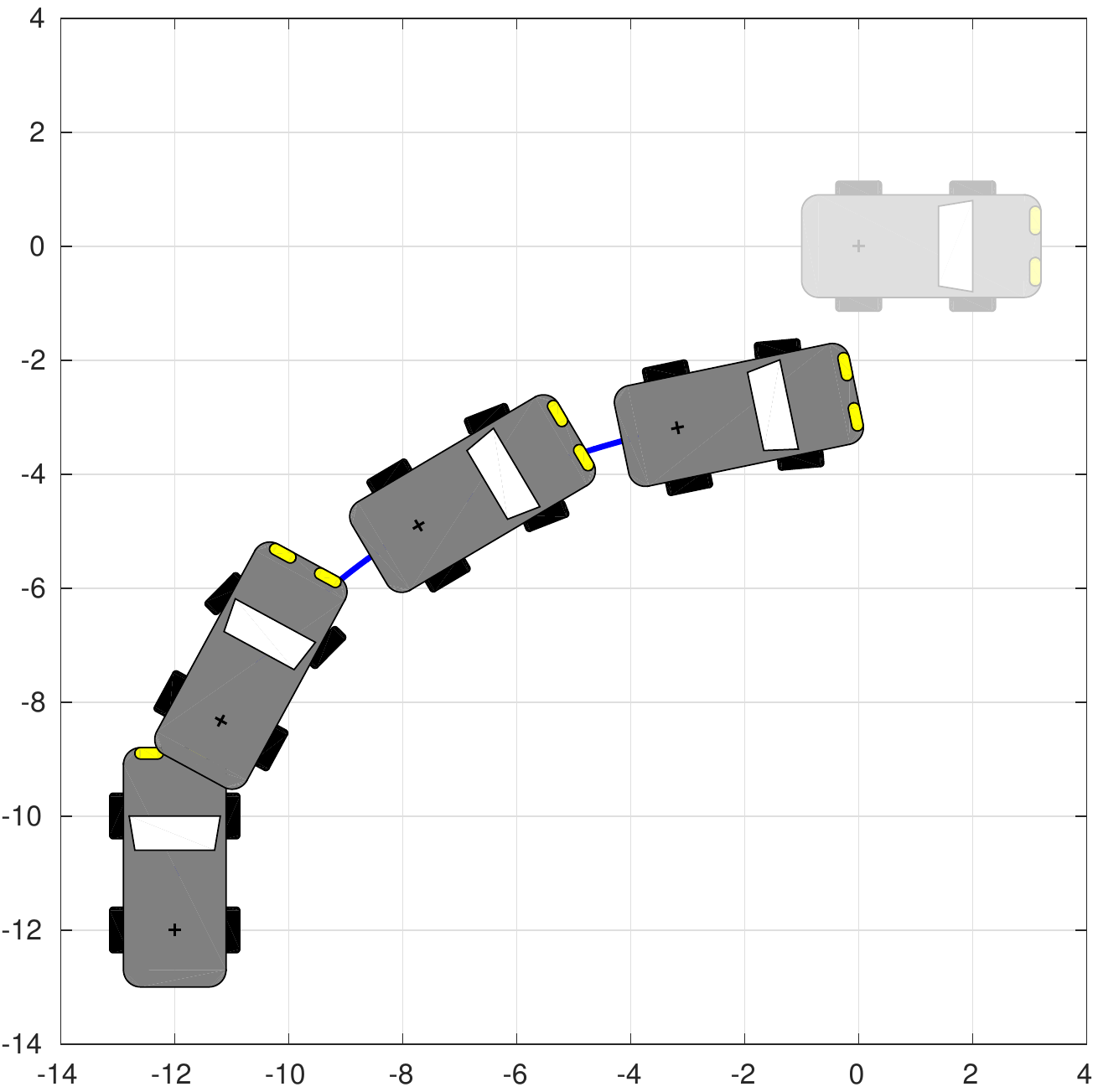}}&
    \fbox{\includegraphics[width=3.2cm]{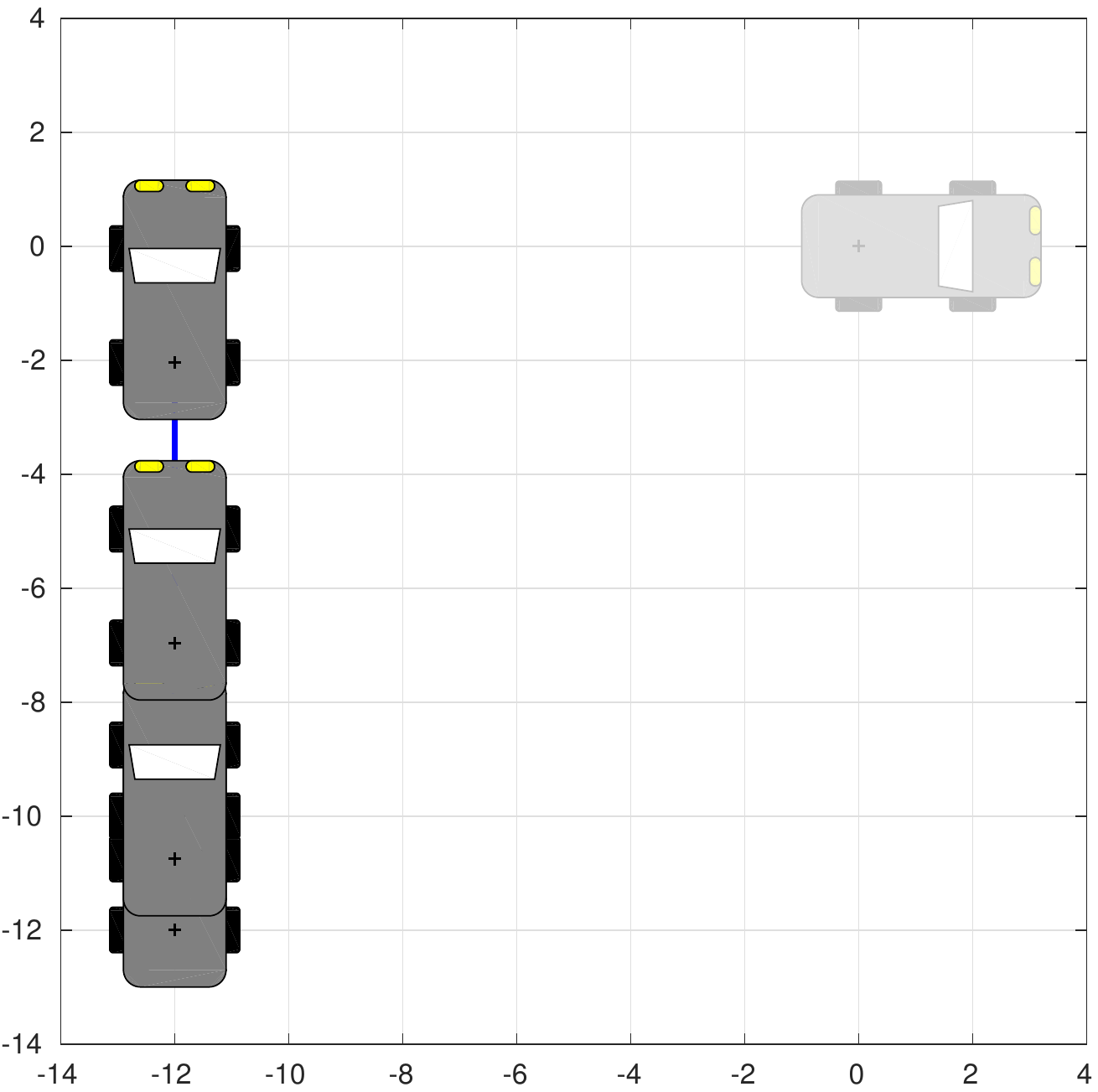}}&
    \fbox{\includegraphics[width=3.2cm]{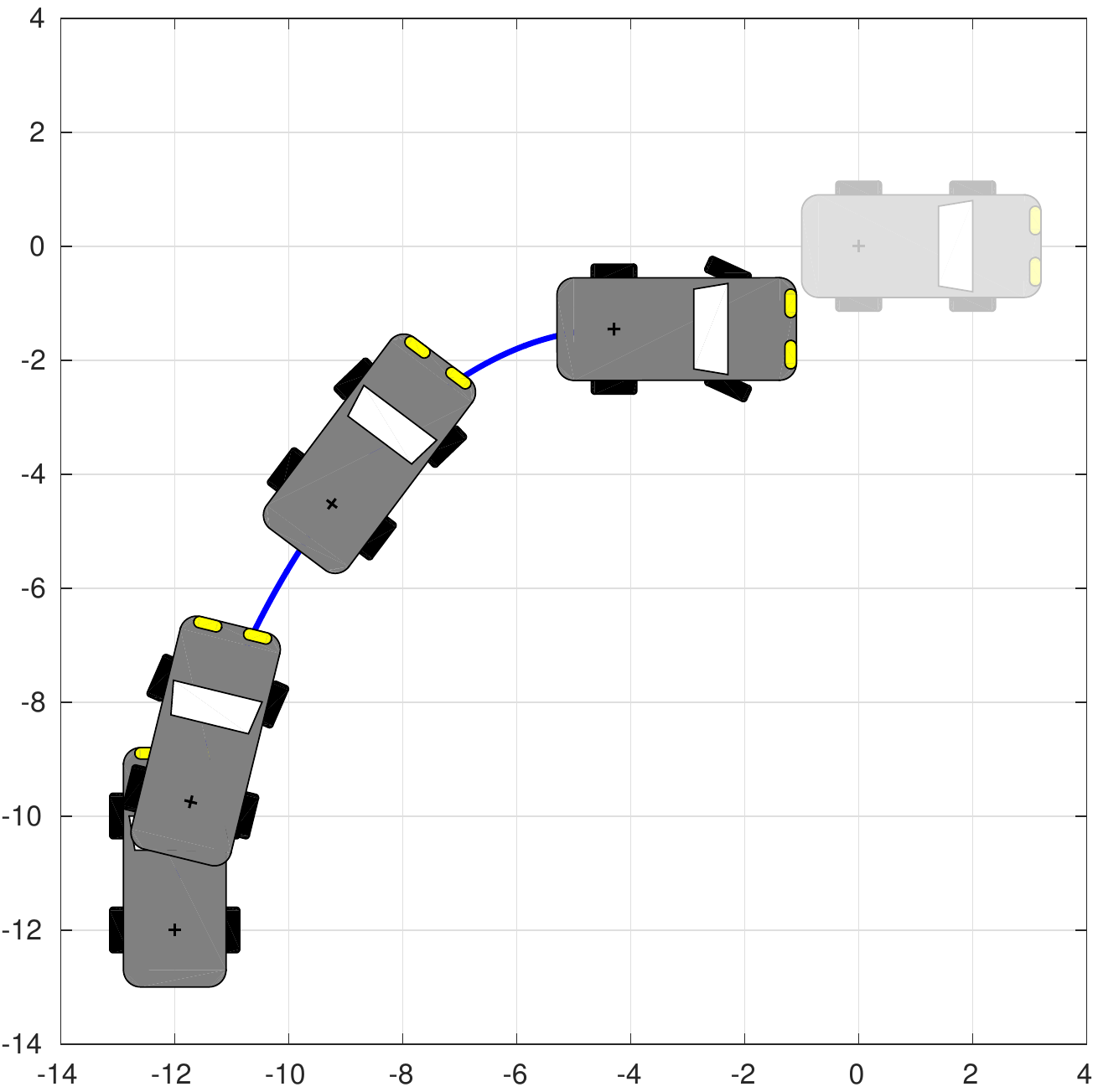}}&
    \fbox{\includegraphics[width=3.2cm]{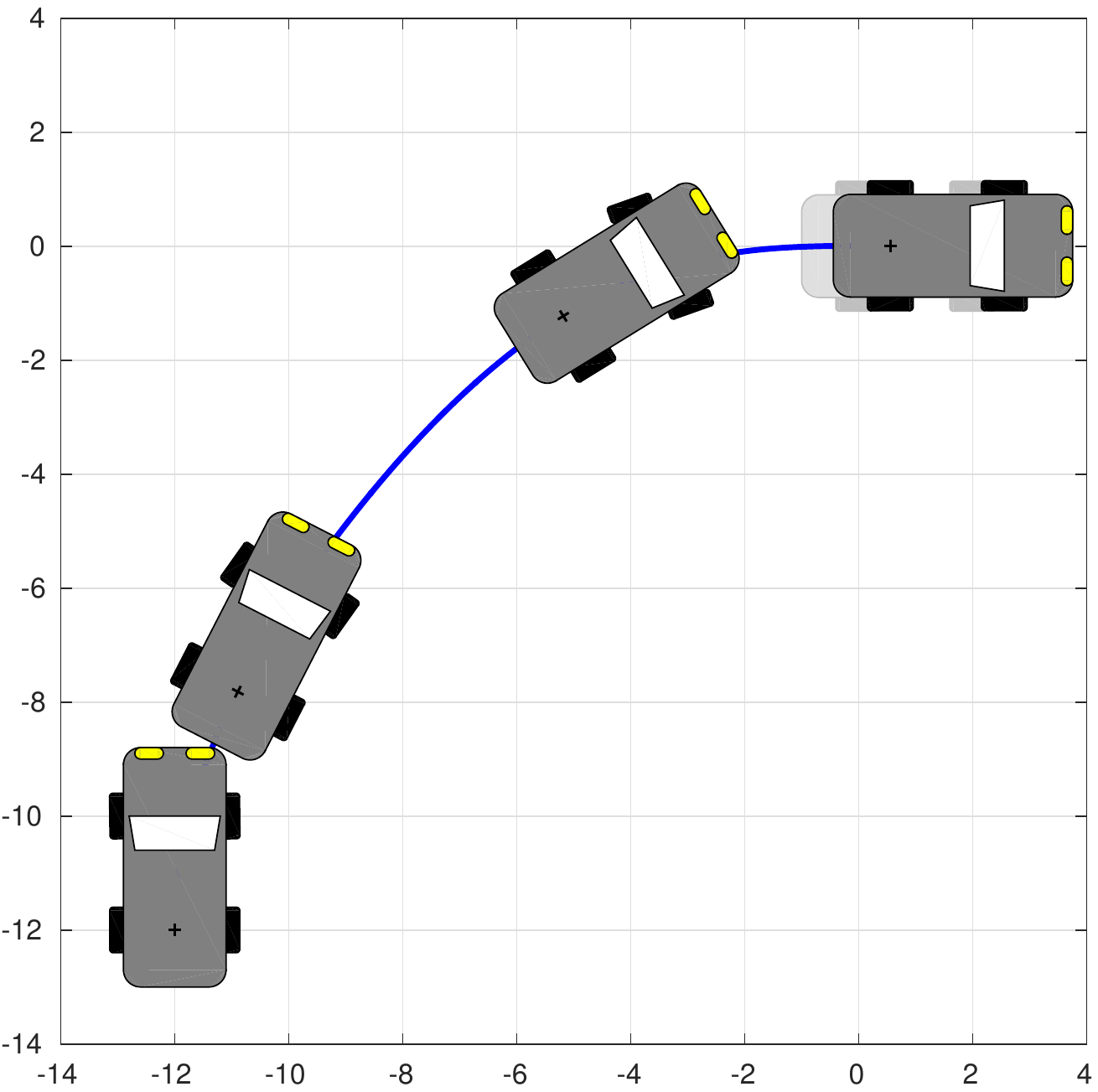}}\\
    \fbox{\includegraphics[width=3.2cm]{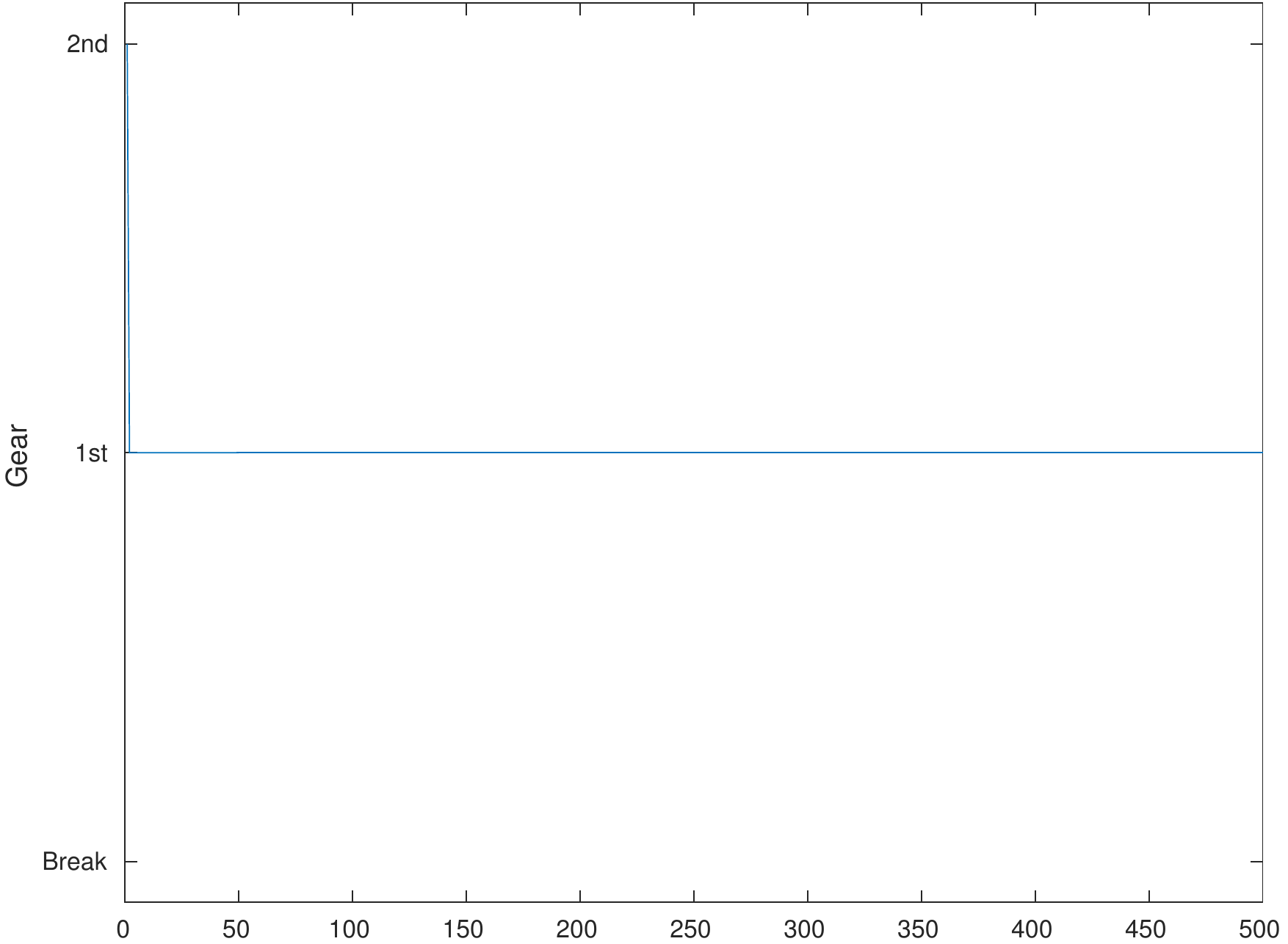}}&
    \fbox{\includegraphics[width=3.2cm]{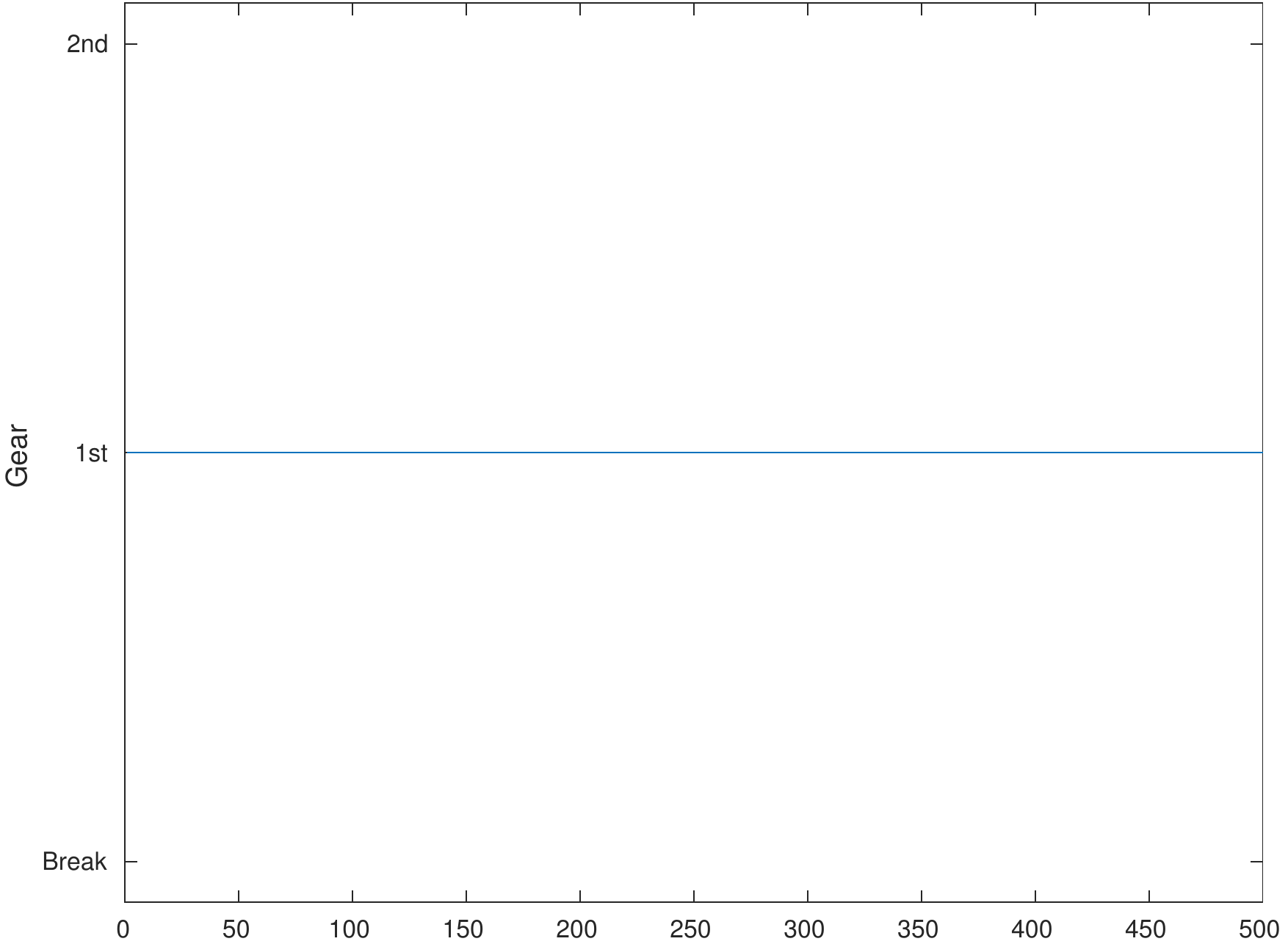}}&
    \fbox{\includegraphics[width=3.2cm]{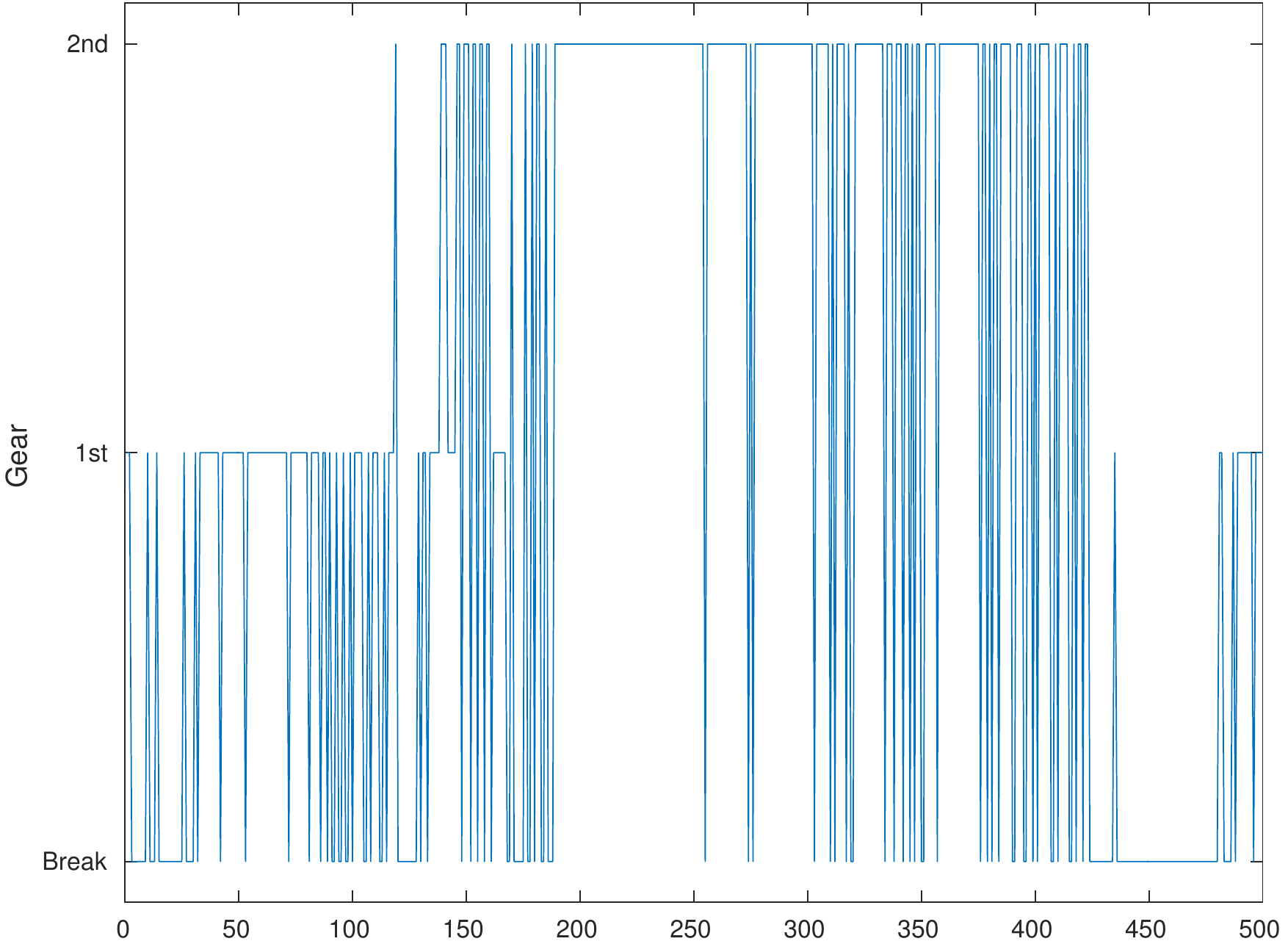}}&
    \fbox{\includegraphics[width=3.2cm]{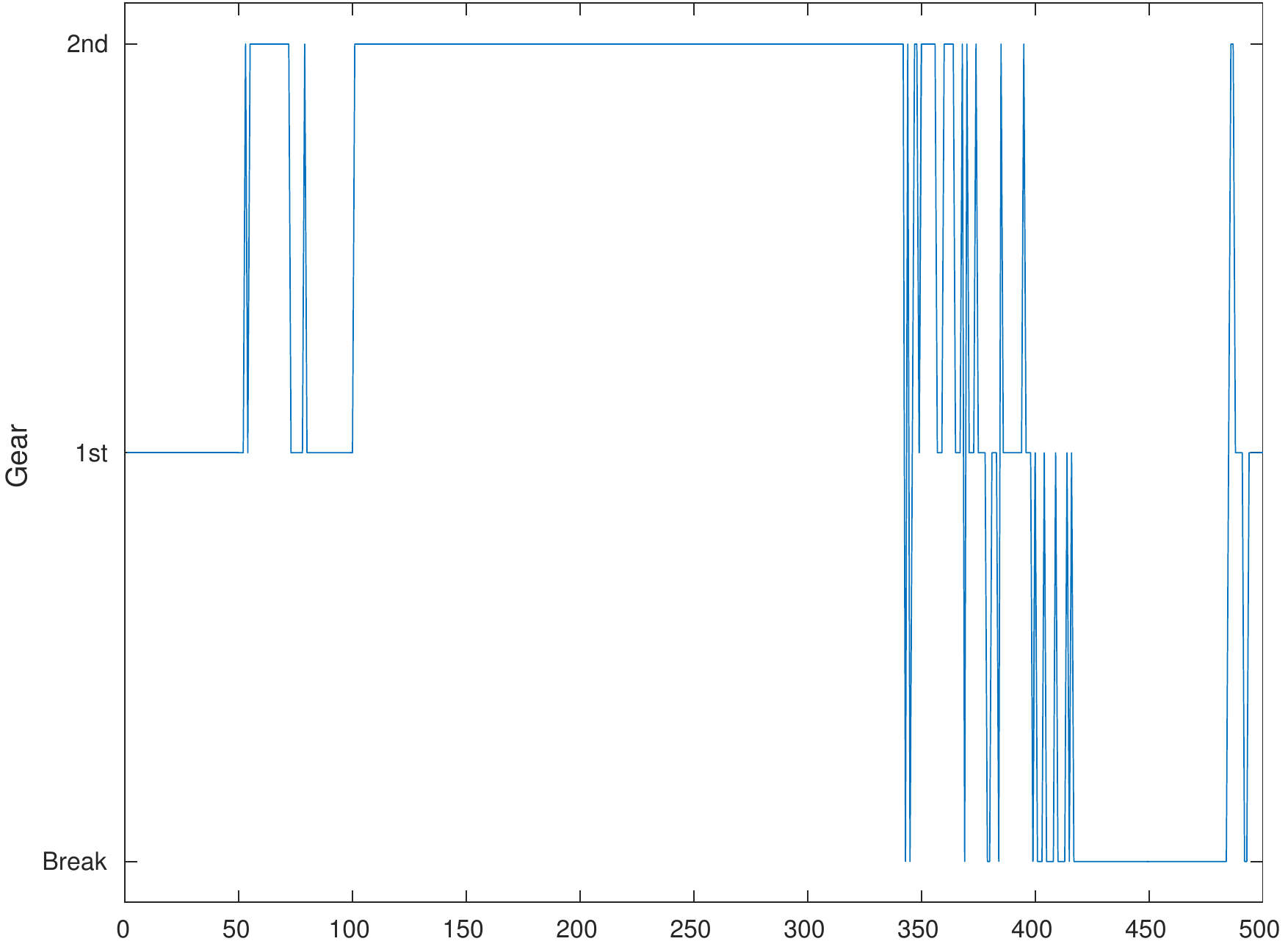}}\\
    \ourfigfont{(a) iLQG} & \ourfigfont{(b) Greedy} &
    \ourfigfont{(c) Interpolate} & \ourfigfont{(d) Mixture} \\
  \end{tabular}
  \caption{Deterministic autonomous car experiment. The goal is to
    drive the car quickly to zero position. The top row shows, for
    each method, the optimized trajectories and the bottom row whether
    the car breaks, uses the 1st gear, or uses the 2nd gear at each
    time step.}
  \label{fig:autonomous_car}
\end{figure*}

For initial controls we used $w_{\textrm{WHEEL}} = 0$, 1st gear,
$acc_{\textrm{CAR}} = 0.1$. Non-optimized code on an Intel i7 CPU took
$0.47$, $0.88$, $0.93$, and $1.03$ seconds per iteration for the
``iLQG'', ``Greedy'', ``Interpolate'', and ``Mixture'' methods,
respectively. Fig.~\ref{fig:results} displays the costs and
Fig.~\ref{fig:autonomous_car} shows the resulting trajectories and
discrete controls. Due to discontinuities, continuous iLQG has
difficulty switching from 1st to 2nd gear and can not achieve maximum
velocity. Since policy improvement starts in DDP from the last time
step and proceeds to the first, greedy iLQG ``Greedy'' does not switch
to 2nd gear. The low cost policy of the proposed hybrid mixture method
``Mixture'' utilizes the 1st gear for fast acceleration, the 2nd gear
for high velocity, and the break for slowing down. To discourage fast
switching one could add a gear/break switching cost.

In addition, we started continuous iLQG from the 2nd gear. Note that
in practice starting from 2nd gear can yield high clutch
wear. ``iLQG'' improved from $9.05$ to $6.96$ total cost while
``Mixture'' was still better with $6.34$ when starting from the 1st
gear, due to ``iLQG'' relying only on the 2nd gear for acceleration
instead of strong 1st gear initial acceleration.


\subsection{Pushing an object}
\label{sec:pushing_unknown_object}

We will now describe the pushing task where the goal is to push an
unknown object into a predefined goal-zone. 

\textbf{State.} We define the state as
\begin{equation}
  \vect{x} = (\vect{x}^C, w, \vect{x}^{CF}, \mu_c, c) \;,
\end{equation}
where $\vect{x}^C = (x^C, y^C)$ denotes the center coordinates and $w$
the rotation angle of the object. $\vect{x}^{CF} = (x^{CF}, y^{CF})$
denotes the center of friction (CF) coordinates relative w.r.t.\
$\vect{x}^C$. $\mu_c$ denotes the friction between the end effector
and the object and $c$ the distribution friction coefficient between
object and supporting surface
\cite{lynch92}. We assume that object edge locations
w.r.t.\ $\vect{x}^C$ are fully observable.


\textbf{Control action.} When pushing we keep the speed of the robot
hand constant while using sufficient force to move the hand. The
discrete control consists of $e$, the discrete edge of the object to
push. The continuous control is
\begin{equation}
  \vect{u} = (u^e, \alpha^p, v),
\end{equation}
where $u^e$, $0 \le u^e \le 1$ is the continuous contact point
location along the edge, $\alpha^p$, $-0.35 \pi \le \alpha_p \le 0.35
\pi$ is the pushing angle w.r.t.\ the normal unit vector at the
contact point w.r.t.\ the edge, and $v$, $0.01 \le v \le 3$ is finger
velocity. It is possible to parameterize $e$ and $u^e$ into a single
continuous control. We do this for the continuous control version of
iLQG. The continuous control version may not be able to jump easily
from one discrete edge to another because of the dynamics
discontinuity at the corners and because of potentially missing the
box when box pose is uncertain.
When pushing the box the finger of the robot can slide along the
pushed edge. Combining sliding and non-sliding dynamics we get the
pushing dynamics as described in~\cite{lynch92}.

\textbf{Observations.} At each time step the robot makes an
observation about the pose of the object. The observation function in
Equation~\ref{eq:observation} is then $\vect{h}[\vect{x}_t] =
(\vect{x}^C, w)$.

\textbf{Cost function.} Our cost function penalizes the robot for not
pushing the object into the target pose by $20 \phi(x^C) + 20
\phi(y^C) + 2 \phi(w)$; at each time step penalizes the distance from
target location by $0.01 (\phi(x^C, 0.1) + \phi(y^C, 0.1))$ and
controls by $10^{-6} ((\alpha^p)^2 + v^2)$; penalizes the robot for
final uncertainty by the sum of all state variances; penalizes for
getting too close to an obstacle at position $\vect{x}^o$ by $- 0.1
\log \Phi ((\vect{x}^o - \vect{x}^C)^T (\vect{x}^o - \vect{x}^C) - 0.5
\sqrt{2})$, where $\Phi(\cdot)$ denotes the Gaussian CDF. Finally, to
avoid missing the object, we penalize pushing too close to a corner
relative to the object rotation variance $\sigma^2_w$ by $0.1 \exp(10
(u^e - \cos(\min (3\sigma^2_w, 0.5 \pi)))) + 0.1 \exp(10 (1 -
\cos(\min (3 \sigma^2_w, 0.5 \pi)) - u^e))$.

In the box pushing experiment, the goal is to push the box to the
target location at zero position. Position $(1,1)$ contains a soft
obstacle. We initialize controls to push the bottom edge $e=0$ and
$u^e = 0.5$, $\alpha^p = 0$, $v = 1$, and, for ``Mixture'' $p_e = 1 -
10^{-10}$. In the fully observable version the robot's planned pushing
location and angle correspond to the real ones. In the partially
observable POMDP problem ``Box POMDP'' the controls are w.r.t.\ the
planned pose and not the actual pose, and the robot may miss the box
resulting in the box not moving. The initial standard deviation (SD)
for the xy-coordinates is $0.01$ and for the rotation angle $0.1$. In
``Box POMDP'', the friction parameters are known. In the ``Box
unknown'' experiments, the coordinates of the center of friction are
unknown and have initially a SD of $0.2$ and in ``Box all unknown''
also the friction parameters $\mu_c$ and $c$ are not known and have
initially a SD of $0.2$. At each time step the robot gets a noisy
observation about the box position with SD $0.0001$ and angle of the
box with SD $0.033$. The SD of dynamics noise for xy-coordinates and
rotation is $0.01$. Friction parameters $\mu_c$ and $c$ had a mean of
$1$.

For evaluation we sampled friction parameters for ``Box unknown'' and
``Box all unknown''. Moreover, in order to test a variety of different
centers of friction (CFs) we selected CFs uniformly between the box
left bottom coordinates $(0.2, 0.2)$ and the top right coordinates
$(0.8, 0.8)$ corresponding to sampling from a uniform
distribution. For the deterministic ``Box'' we ``sampled'' 52
different CFs and for each of the stochastic ``Box POMDP'', ``Box
unknown'', and ``Box all unknown'' problems 12 different CFs. In the
stochastic problems we averaged costs for each CF over 20 sampled
trajectories.

\begin{figure*}
  \centering
  \vspace{0.4em}
  \setlength{\tabcolsep}{1pt} 
  \begin{tabular}{lcccc}
    \raisebox{7\height}{\ourfigfont{iLQG}} &
    \fbox{\includegraphics[width=3.2cm]{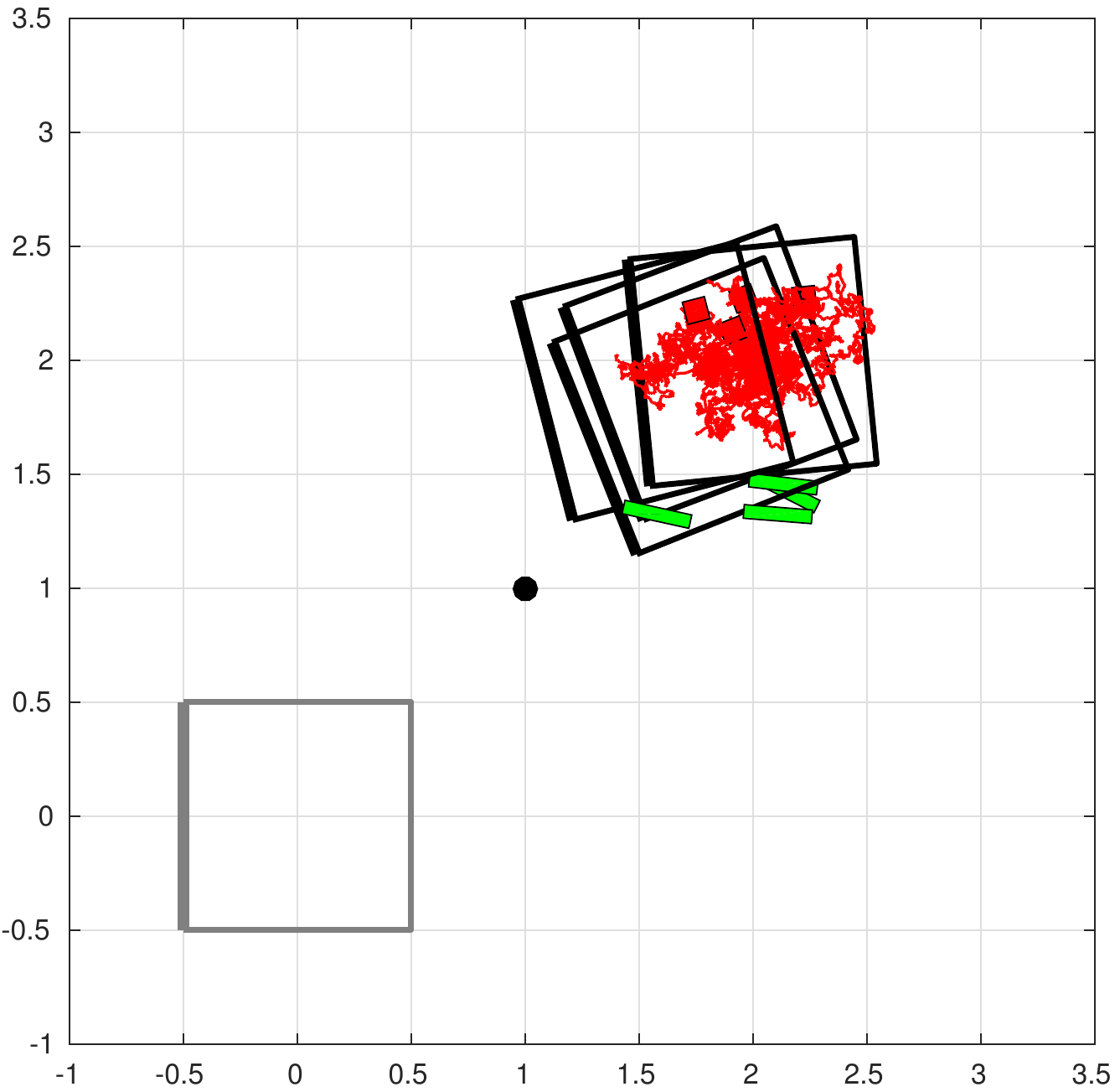}}&
    \fbox{\includegraphics[width=3.2cm]{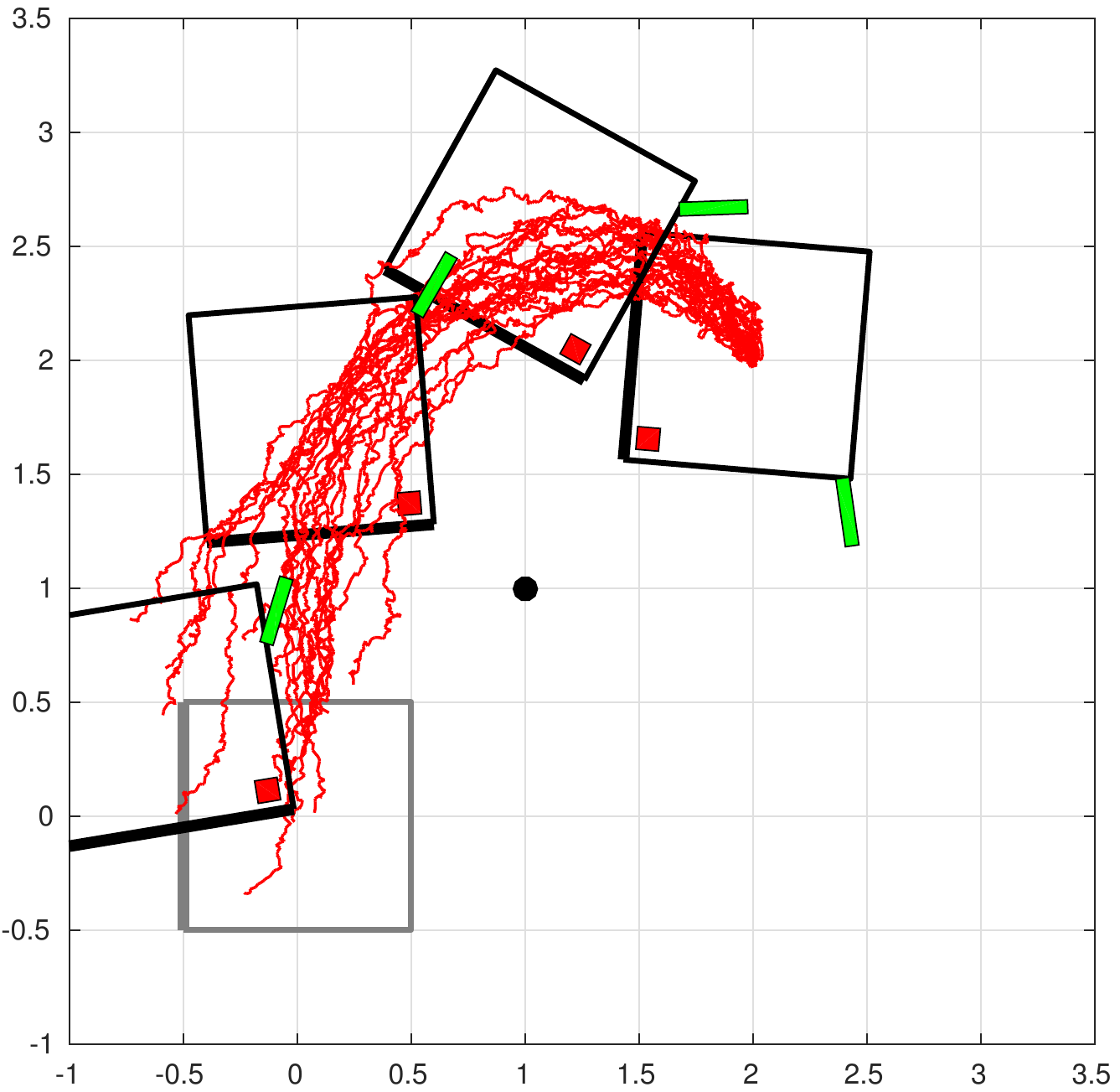}}&
    \fbox{\includegraphics[width=3.2cm]{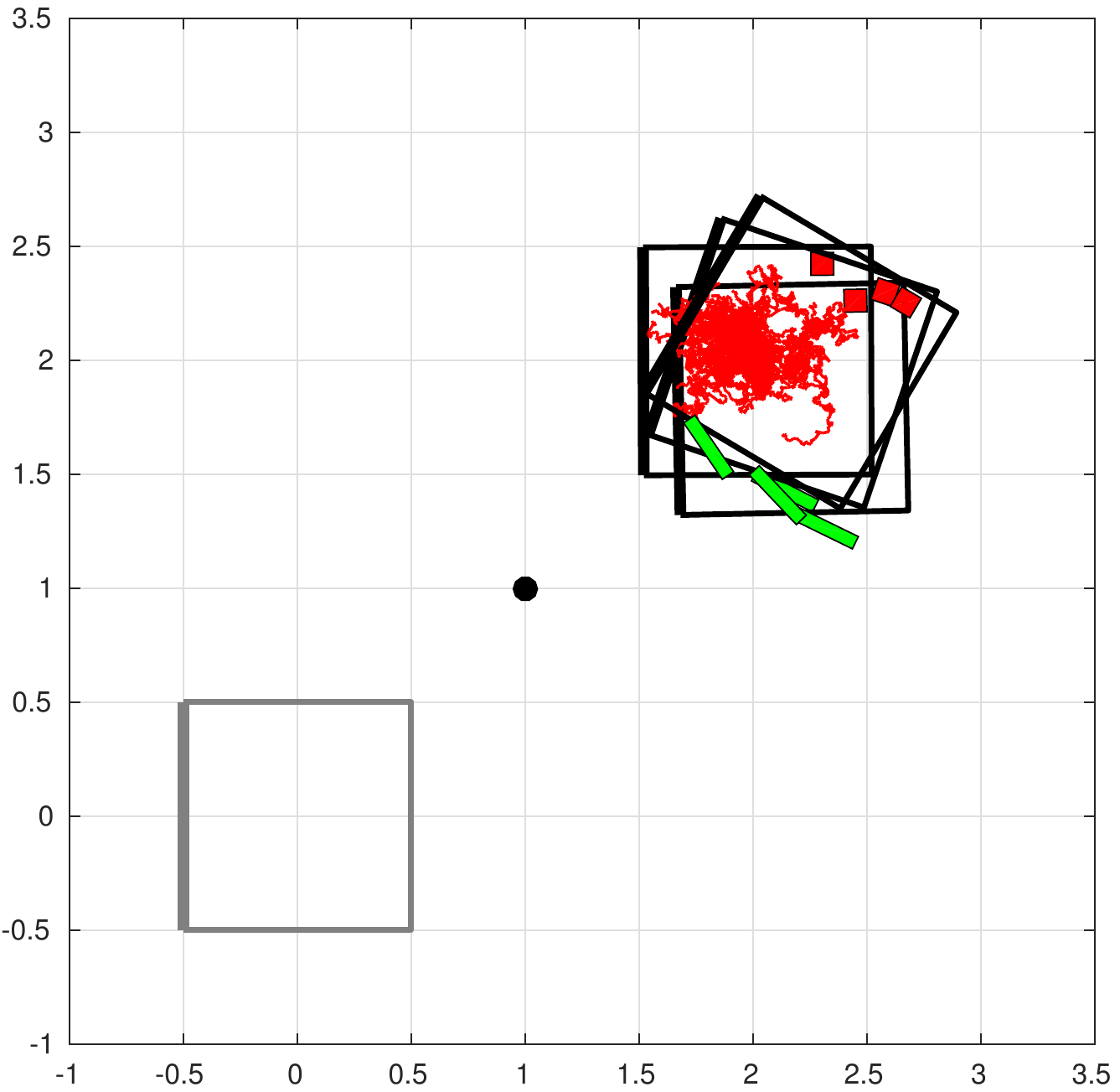}}&
    \fbox{\includegraphics[width=3.2cm]{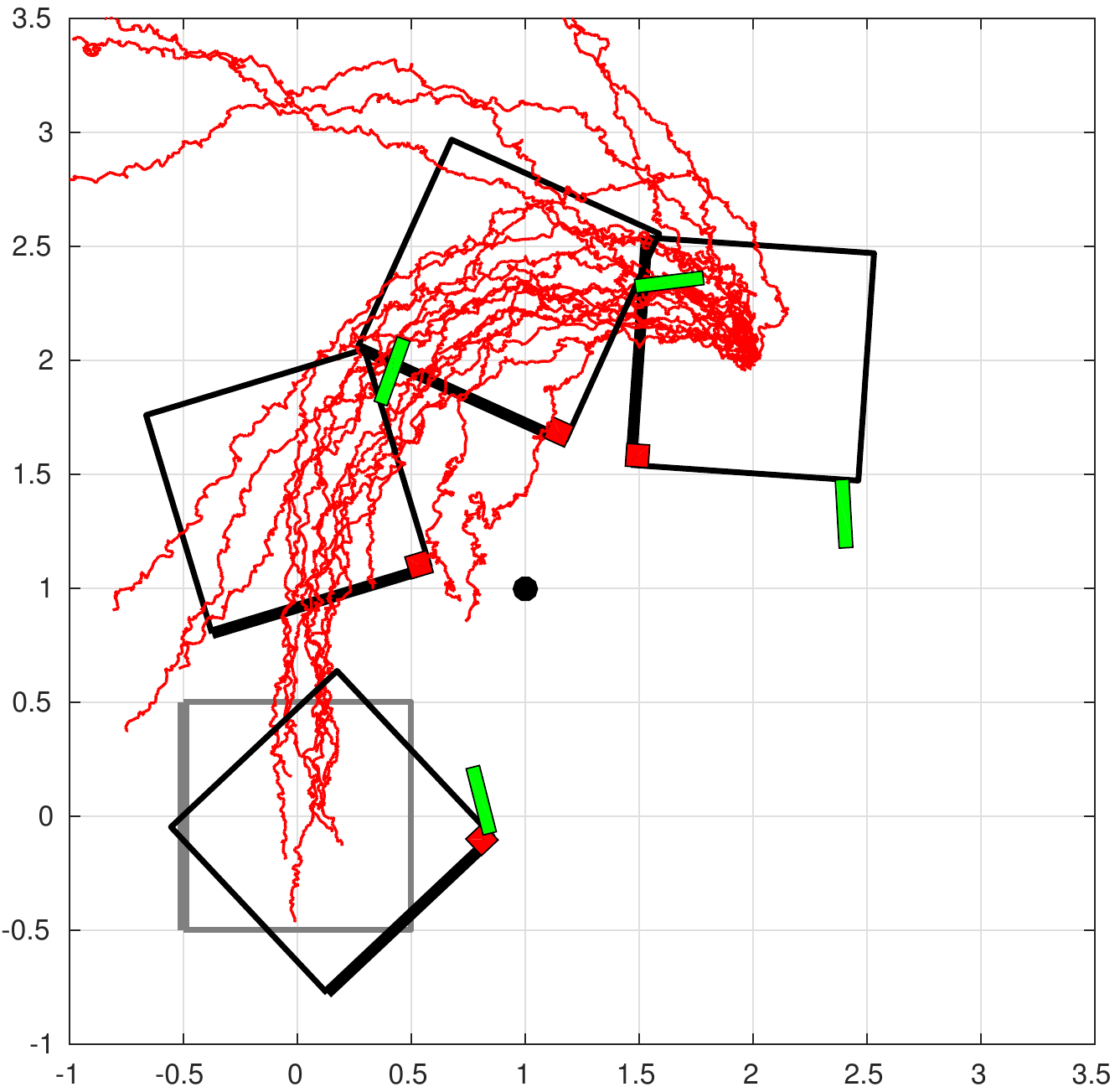}}\\[3pt]
    \raisebox{7\height}{\ourfigfont{Mixture}} &
    \fbox{\includegraphics[width=3.2cm]{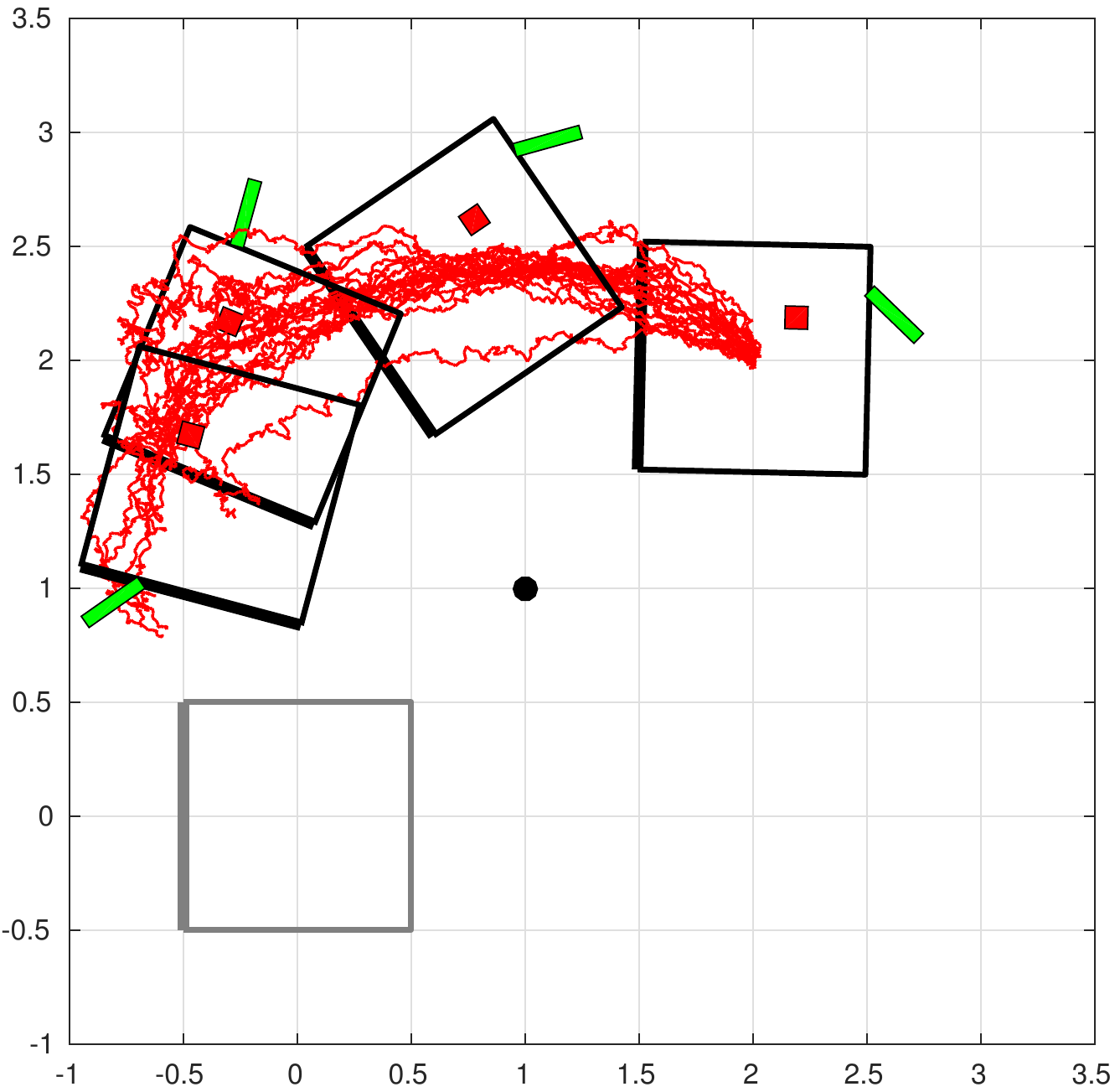}}&
    \fbox{\includegraphics[width=3.2cm]{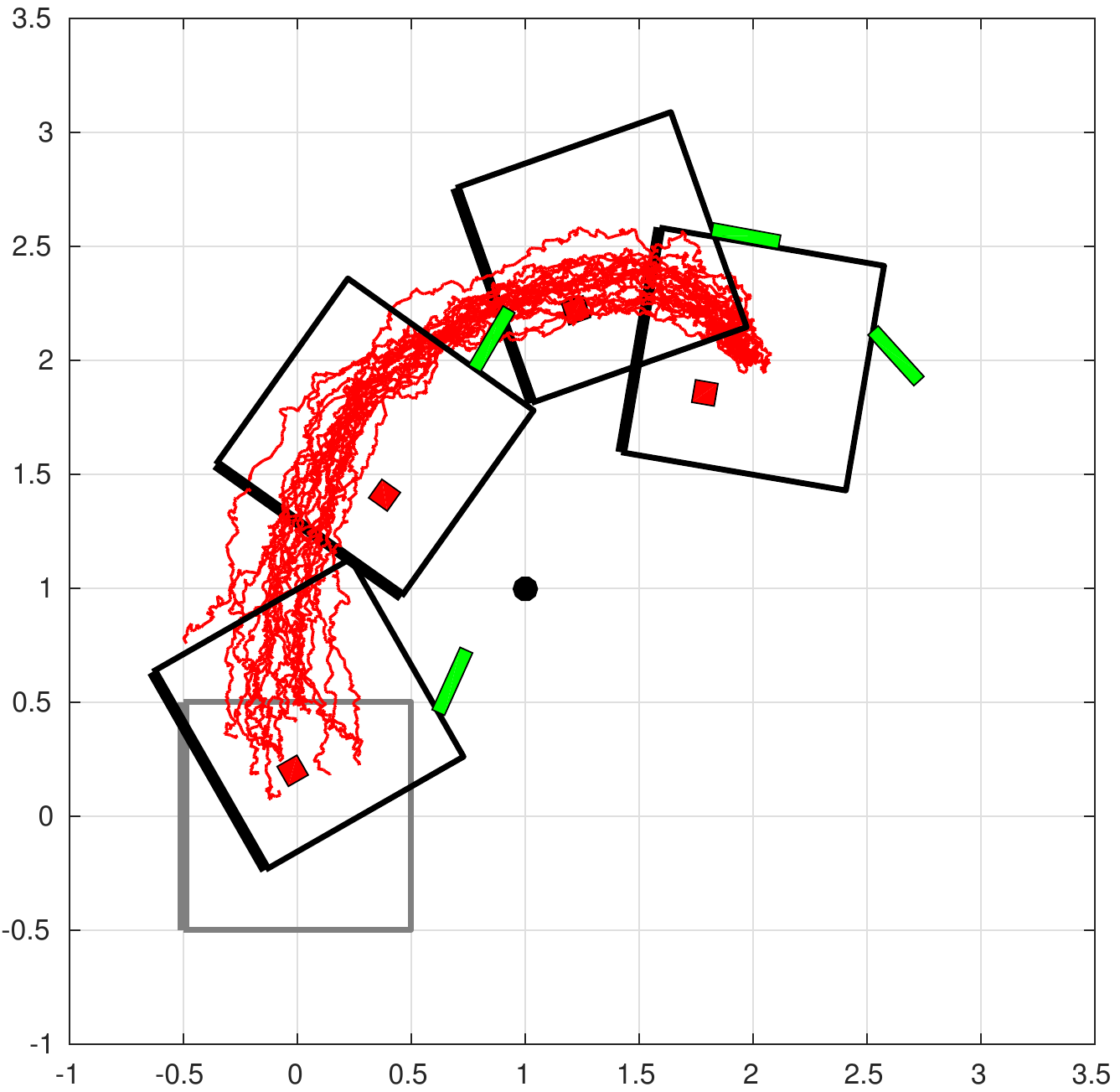}}&
    \fbox{\includegraphics[width=3.2cm]{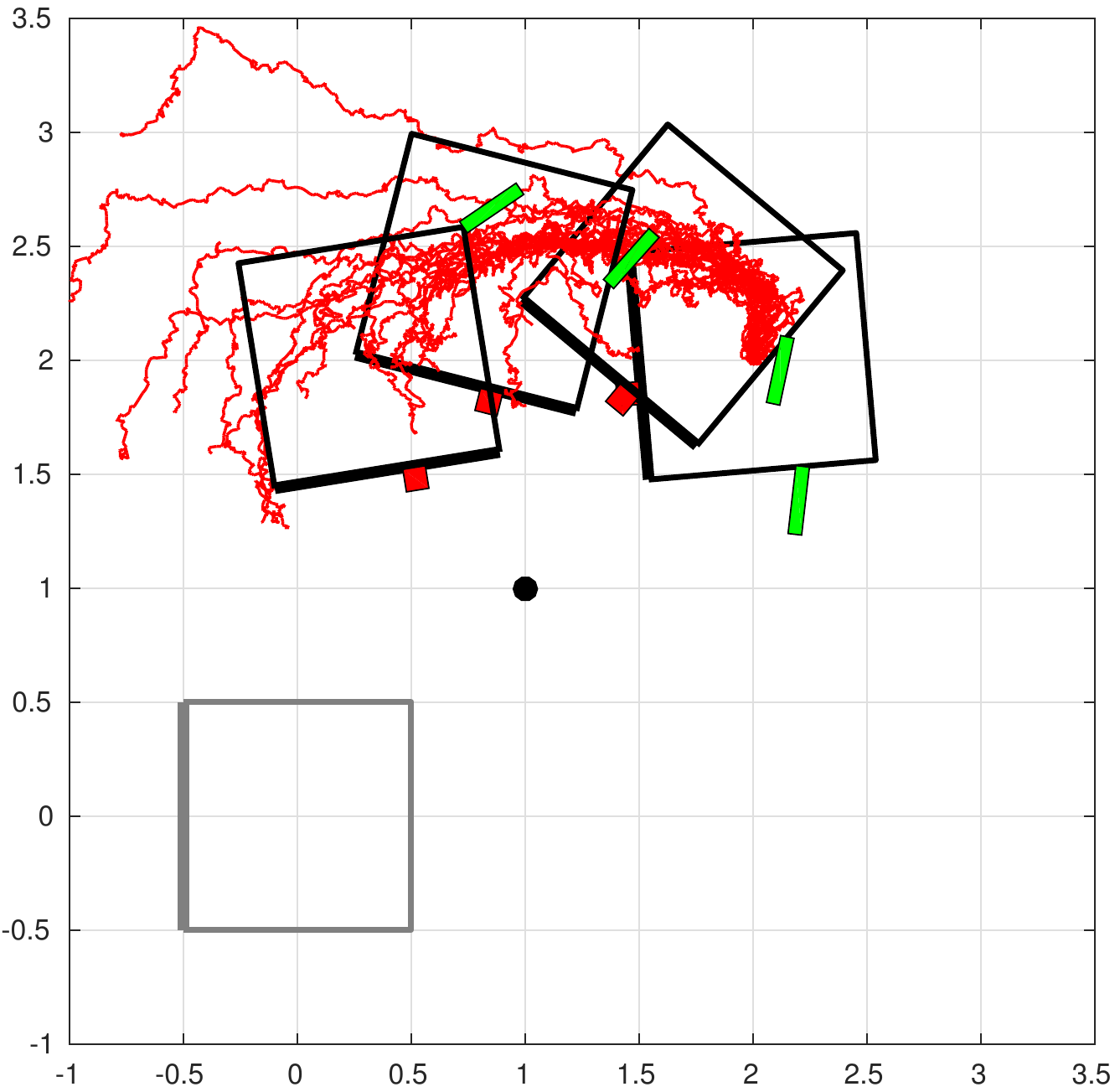}}&
    \fbox{\includegraphics[width=3.2cm]{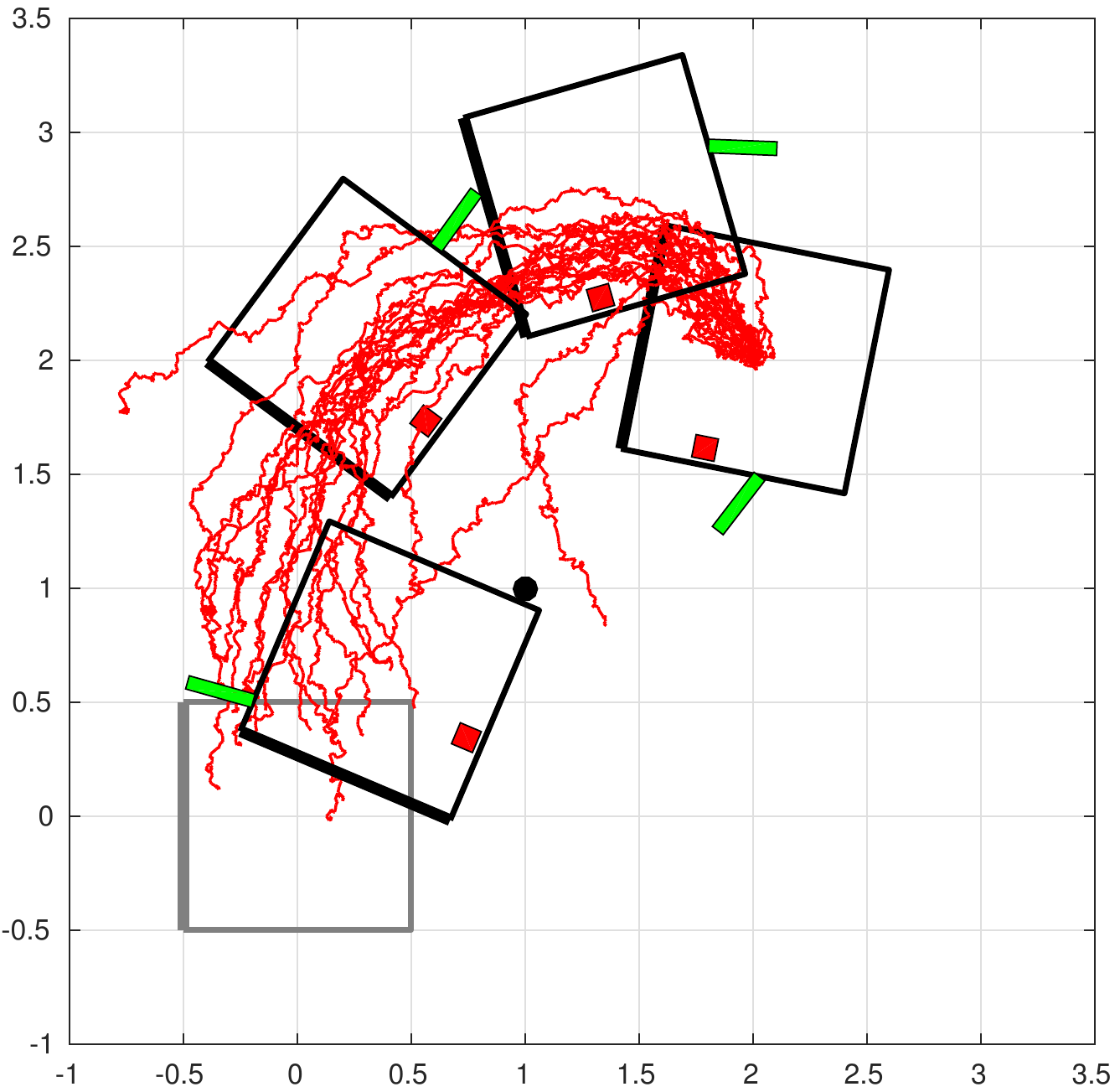}}\\
    & \ourfigfont{(a) Worst} & \ourfigfont{(b) Best} & \ourfigfont{(c) Worst} & \ourfigfont{(d) Best} \\
    & \multicolumn{2}{c}{\ourfigfont{Box POMDP}} & \multicolumn{2}{c}{\ourfigfont{Box unknown}} \\
  \end{tabular}
  \caption{Uncertain box pushing. The goal is to push the box using
    the green finger to zero position while avoiding the obstacle at
    $(1,1)$. For (a) and (b) the box position and rotation are
    uncertain and partially observable, but for (c) and (d) also the
    red center of friction is uncertain. (a) and (c) show worst
    performance and (b) and (d) the best. For one of the sampled red
    trajectories ($20$ in each plot) we display four boxes and finger
    configurations distributed evenly over time. For rotation
    visualization one box edge is thicker than the others.}
  \vspace{-1em}
  \label{fig:box_pushing}
\end{figure*}

Fig.~\ref{fig:results} shows the cost means and standard errors over
the CFs. The ``iLQG'', ``Greedy'', ``Interpolate'', and ``Mixture''
methods, on the ``Box''/``Box POMDP''/``Box unknown''/``Box all
unknown'' problems took $0.86/3.03/4.32/8.61$, $1.36/3.53/4.55/8.74$,
$1.37/3.75/5.28/10.95$, and $1.59/5.80/8.60/18.06$ seconds per
iteration, respectively. ``Mixture'' performs best. As expected higher
uncertainties decrease performance of ``Mixture''. ``Interpolate'',
``Greedy'', and ``iLQG'' seem to have systematic problems in all
setups with high uncertainty. The heuristics of ``Interpolate'' and
``Greedy'' do not always work and iLQG gets stuck in local
optima. Fig.~\ref{fig:box_pushing} shows high and low cost examples
for both the ``Box POMDP'' and ``Box unknown'' problems for ``iLQG''
and our ``Mixture'' method. In the worst case, iLQG can not escape
local optimums: the cost for potentially missing the box prevents
switching pushing sides. In the POMDP problem, for a suitable center
of friction, iLQG computes a good policy but in the unknown POMDP
problem has even in the best case run away trajectories.



\section{CONCLUSIONS}
\label{sec:conclusions}
We presented a novel DDP approach with linear feedback control for
hybrid control of trajectories under uncertainty. The experiments
indicate that our approach is useful, especially in POMDP problems. In
the future, we plan on applying the approach to a real robot. We
may use online replanning to improve the results further.

\bibliographystyle{IEEEtran}
\bibliography{root}

\begin{thebibliography}{10}
\providecommand{\url}[1]{#1}
\csname url@rmstyle\endcsname
\providecommand{\newblock}{\relax}
\providecommand{\bibinfo}[2]{#2}
\providecommand\BIBentrySTDinterwordspacing{\spaceskip=0pt\relax}
\providecommand\BIBentryALTinterwordstretchfactor{4}
\providecommand\BIBentryALTinterwordspacing{\spaceskip=\fontdimen2\font plus
\BIBentryALTinterwordstretchfactor\fontdimen3\font minus
  \fontdimen4\font\relax}
\providecommand\BIBforeignlanguage[2]{{%
\expandafter\ifx\csname l@#1\endcsname\relax
\typeout{** WARNING: IEEEtran.bst: No hyphenation pattern has been}%
\typeout{** loaded for the language `#1'. Using the pattern for}%
\typeout{** the default language instead.}%
\else
\language=\csname l@#1\endcsname
\fi
#2}}

\bibitem{kirches09}
C.~Kirches, S.~Sager, H.~G. Bock, and J.~P. Schl{\"o}der, ``Time-optimal
  control of automobile test drives with gear shifts,'' \emph{Optimal Control
  Applications and Methods}, vol.~31, no.~2, pp. 137--153, 2010.

\bibitem{dogar10}
M.~Dogar and S.~Srinivasa, ``Push-grasping with dexterous hands: Mechanics and
  a method,'' in \emph{Proceedings of the IEEE/RSJ International Conference on
  Intelligent Robots and Systems (IROS)}, 2010.

\bibitem{dogar12}
------, ``A planning framework for non-prehensile manipulation under clutter
  and uncertainty,'' \emph{Autonomous Robots}, vol.~33, no.~3, pp. 217--236,
  2012.

\bibitem{koval14}
M.~Koval, N.~Pollard, and S.~Srinivasa, ``Pre- and post-contact policy
  decomposition for planar contact manipulation under uncertainty,'' in
  \emph{Proceedings of Robotics: Science and Systems (R:SS)}, 2014.

\bibitem{kawajiri08}
Y.~Kawajiri and L.~T. Biegler, ``Large scale optimization strategies for zone
  configuration of simulated moving beds,'' \emph{Computers \& Chemical
  Engineering}, vol.~32, no.~1, pp. 135--144, 2008.

\bibitem{branicky98}
M.~S. Branicky, V.~S. Borkar, and S.~K. Mitter, ``A unified framework for
  hybrid control: Model and optimal control theory,'' \emph{IEEE Transactions
  on Automatic Control}, vol.~43, no.~1, pp. 31--45, 1998.

\bibitem{bemporad00}
A.~Bemporad, G.~Ferrari-Trecate, and M.~Morari, ``Observability and
  controllability of piecewise affine and hybrid systems,'' \emph{IEEE
  Transactions on Automatic Control}, vol.~45, no.~10, pp. 1864--1876, 2000.

\bibitem{sager05}
S.~Sager, \emph{Numerical methods for mixed-integer optimal control
  problems}.\hskip 1em plus 0.5em minus 0.4em\relax Der andere Verlag
  T{\"o}nning, L{\"u}beck, Marburg, 2005.

\bibitem{nandola08}
N.~N. Nandola and S.~Bhartiya, ``A multiple model approach for predictive
  control of nonlinear hybrid systems,'' \emph{Journal of process control},
  vol.~18, no.~2, pp. 131--148, 2008.

\bibitem{azhmyakov09}
V.~Azhmyakov, R.~Galvan-Guerra, and M.~Egerstedt, ``Hybrid lq-optimization
  using dynamic programming,'' in \emph{Proceedings of the American Control
  Conference}.\hskip 1em plus 0.5em minus 0.4em\relax IEEE, 2009, pp.
  3617--3623.

\bibitem{zhu15}
F.~Zhu and P.~J. Antsaklis, ``Optimal control of hybrid switched systems: A
  brief survey,'' \emph{Discrete Event Dynamic Systems}, vol.~25, no.~3, pp.
  345--364, 2015.

\bibitem{tassa14}
Y.~Tassa, N.~Mansard, and E.~Todorov, ``Control-limited differential dynamic
  programming,'' in \emph{Proceedings of the IEEE International Conference on
  Robotics and Automation (ICRA)}.\hskip 1em plus 0.5em minus 0.4em\relax IEEE,
  2014, pp. 1168--1175.

\bibitem{riedinger99}
P.~Riedinger, F.~Kratz, C.~Iung, and C.~Zanne, ``Linear quadratic optimization
  for hybrid systems,'' in \emph{IEEE Conference on Decision and Control},
  vol.~3.\hskip 1em plus 0.5em minus 0.4em\relax IEEE, 1999, pp. 3059--3064.

\bibitem{vonstryk92}
O.~Von~Stryk and R.~Bulirsch, ``Direct and indirect methods for trajectory
  optimization,'' \emph{Annals of operations research}, vol.~37, no.~1, pp.
  357--373, 1992.

\bibitem{platt10}
R.~Platt~Jr, R.~Tedrake, L.~Kaelbling, and T.~Lozano-Perez, ``Belief space
  planning assuming maximum likelihood observations,'' in \emph{Robotics:
  Science and Systems (RSS)}, 2010.

\bibitem{van12}
J.~van~den Berg, S.~Patil, and R.~Alterovitz, ``{Efficient Approximate Value
  Iteration for Continuous Gaussian POMDPs},'' in \emph{Proceedings of the AAAI
  Conference on Artificial Intelligence}.\hskip 1em plus 0.5em minus
  0.4em\relax AAAI Press, 2012.

\bibitem{patil15}
S.~Patil, G.~Kahn, M.~Laskey, J.~Schulman, K.~Goldberg, and P.~Abbeel,
  ``Scaling up gaussian belief space planning through covariance-free
  trajectory optimization and automatic differentiation,'' in \emph{Algorithmic
  Foundations of Robotics XI}.\hskip 1em plus 0.5em minus 0.4em\relax Springer,
  2015, pp. 515--533.

\bibitem{lincoln02}
B.~Lincoln and B.~Bernhardsson, ``Lqr optimization of linear system
  switching,'' \emph{IEEE Transactions on Automatic Control}, vol.~47, no.~10,
  pp. 1701--1705, 2002.

\bibitem{daniel16}
C.~Daniel, H.~van Hoof, J.~Peters, and G.~Neumann, ``Probabilistic inference
  for determining options in reinforcement learning,'' in \emph{Proceedings of
  The European Conference on Machine Learning and Principles and Practice of
  Knowledge Discovery}.\hskip 1em plus 0.5em minus 0.4em\relax Springer, 2016.

\bibitem{lavalle01}
S.~LaValle and J.~Kuffner, ``Rapidly-exploring random trees: Progress and
  prospects,'' \emph{Algorithmic and Computational Robotics: New Directions},
  pp. 293--308, 2001.

\bibitem{branicky02}
M.~S. Branicky and M.~M. Curtiss, ``Nonlinear and hybrid control via rrts,'' in
  \emph{Proc. Intl. Symp. on Mathematical Theory of Networks and Systems}, vol.
  750, 2002.

\bibitem{zito12}
C.~Zito, R.~Stolkin, M.~Kopicki, and J.~L. Wyatt, ``{Two-level RRT planning for
  robotic push manipulation},'' in \emph{Proc. of IEEE/RSJ International
  Conference on Intelligent Robots and Systems (IROS)}.\hskip 1em plus 0.5em
  minus 0.4em\relax IEEE, 2012, pp. 678--685.

\bibitem{egerstedt06}
M.~Egerstedt, S.-i. Azuma, and Y.~Wardi, ``Optimal timing control of switched
  linear systems based on partial information,'' \emph{Nonlinear Analysis:
  Theory, Methods \& Applications}, vol.~65, no.~9, pp. 1736--1750, 2006.

\bibitem{azuma06}
S.-i. Azuma, M.~Egerstedt, and Y.~Wardi, ``Output-based optimal timing control
  of switched systems,'' in \emph{International Workshop on Hybrid Systems:
  Computation and Control}.\hskip 1em plus 0.5em minus 0.4em\relax Springer,
  2006, pp. 64--78.

\bibitem{ross08}
S.~Ross, B.~Chaib-draa, and J.~Pineau, ``{Bayesian reinforcement learning in
  continuous POMDPs with application to robot navigation},'' in
  \emph{Proceedings of the IEEE International Conference on Robotics and
  Automation (ICRA)}.\hskip 1em plus 0.5em minus 0.4em\relax IEEE, 2008, pp.
  2845--2851.

\bibitem{dallaire09}
P.~Dallaire, C.~Besse, S.~Ross, and B.~Chaib-draa, ``{Bayesian reinforcement
  learning in continuous POMDPs with Gaussian Processes},'' in
  \emph{Proceedings of the IEEE/RSJ International Conference on Intelligent
  Robots and Systems (IROS)}.\hskip 1em plus 0.5em minus 0.4em\relax IEEE,
  2009, pp. 2604--2609.

\bibitem{bai14}
H.~Bai, D.~Hsu, and W.~S. Lee, ``{Integrated perception and planning in the
  continuous space: A POMDP approach},'' \emph{The International Journal of
  Robotics Research}, vol.~33, no.~9, pp. 1288--1302, 2014.

\bibitem{agha14}
A.-A. Agha-Mohammadi, S.~Chakravorty, and N.~M. Amato, ``{FIRM: Sampling-based
  feedback motion planning under motion uncertainty and imperfect
  measurements},'' \emph{The International Journal of Robotics Research},
  vol.~33, no.~2, pp. 268--304, 2014.

\bibitem{kopicki16}
M.~Kopicki, S.~Zurek, R.~Stolkin, T.~Moerwald, and J.~L. Wyatt, ``Learning
  modular and transferable forward models of the motions of push manipulated
  objects,'' \emph{Autonomous Robots}, pp. 1--22, 2016.

\bibitem{lynch92}
K.~M. Lynch, H.~Maekawa, and K.~Tanie, ``Manipulation and active sensing by
  pushing using tactile feedback.'' in \emph{Proc. of IEEE/RSJ International
  Conference on Intelligent Robots and Systems (IROS)}.\hskip 1em plus 0.5em
  minus 0.4em\relax IEEE, 1992, pp. 416--421.

\bibitem{koval16}
M.~C. Koval, N.~S. Pollard, and S.~S. Srinivasa, ``Pre-and post-contact policy
  decomposition for planar contact manipulation under uncertainty,'' \emph{The
  International Journal of Robotics Research}, vol.~35, no. 1-3, pp. 244--264,
  2016.

\bibitem{webb14}
D.~J. Webb, K.~L. Crandall, and J.~van~den Berg, ``{Online Parameter Estimation
  via Real-Time Replanning of Continuous Gaussian POMDPs},'' in
  \emph{Proceedings of the IEEE International Conference on Robotics and
  Automation (ICRA)}.\hskip 1em plus 0.5em minus 0.4em\relax IEEE, 2014.

\bibitem{mayne66}
D.~Mayne, ``A second-order gradient method for determining optimal trajectories
  of non-linear discrete-time systems,'' \emph{International Journal of
  Control}, vol.~3, no.~1, pp. 85--95, 1966.

\bibitem{jacobson70}
D.~Jacobson and D.~Mayne, ``Differential dynamic programming,'' 1970.

\bibitem{liao92}
L.-z. Liao and C.~A. Shoemaker, ``Advantages of differential dynamic
  programming over newton's method for discrete-time optimal control
  problems,'' Cornell University, Tech. Rep., 1992.

\bibitem{todorov05}
E.~Todorov and W.~Li, ``A generalized iterative lqg method for locally-optimal
  feedback control of constrained nonlinear stochastic systems,'' in
  \emph{Proceedings of the American Control Conference}.\hskip 1em plus 0.5em
  minus 0.4em\relax IEEE, 2005, pp. 300--306.

\end{thebibliography}

\end{document}